\documentclass{template}

\usepackage{lmodern}
\usepackage{xspace}
\usepackage[authoryear,round]{natbib}
\usepackage[table]{xcolor}
\usepackage{wrapfig}
\usepackage{graphicx,amsmath,amssymb}
\usepackage{booktabs}
\usepackage{url}
\usepackage{float}
\usepackage{subcaption}
\usepackage{multirow}
\usepackage{tabularx}
\usepackage{array}
\usepackage{adjustbox}
\usepackage{listings}
\usepackage{tcolorbox}
\usepackage{twemojis}

\definecolor{link}{HTML}{0063BE}
\hypersetup{
  colorlinks=true,
  urlcolor=link,
  linkcolor=black,
  citecolor=link,
  filecolor=black,
  pdfborder={0 0 0}
}

\providecommand{\equationname}{Equation}
\providecommand{\sectionname}{Section}
\providecommand{\appendixname}{Appendix}

\newcommand{\figref}[1]{%
  \figurename~\hyperref[#1]{\textcolor{link}{\ref*{#1}}}%
}
\newcommand{\tabref}[1]{%
  \tablename~\hyperref[#1]{\textcolor{link}{\ref*{#1}}}%
}
\newcommand{\eqrefc}[1]{%
  \equationname~\hyperref[#1]{\textcolor{link}{\ref*{#1}}}%
}
\newcommand{\secref}[1]{%
  \sectionname~\hyperref[#1]{\textcolor{link}{\ref*{#1}}}%
}
\newcommand{\appendixref}[1]{%
  \appendixname~\hyperref[#1]{\textcolor{link}{\ref*{#1}}}%
}

\renewcommand{\paragraph}[1]{\textbf{#1.}}
\newcolumntype{x}[1]{>{\centering\arraybackslash}p{#1pt}}
\newcolumntype{y}[1]{>{\raggedright\arraybackslash}p{#1pt}}
\newcolumntype{z}[1]{>{\raggedleft\arraybackslash}p{#1pt}}
\setlist{nosep}

\newlength\savewidth

\newcommand{\commentone}[1]{}
\newcommand{\commenttwo}[1]{}

\newcommand{\model}{AgentVQA}

\definecolor{domainred}{HTML}{FF8988}
\definecolor{domainorange}{HTML}{FECC81}
\definecolor{domainyellow}{HTML}{F5DC51}
\definecolor{domainblue}{HTML}{6098FF}
\definecolor{domaingreen}{HTML}{77B25D}
\definecolor{domainpurple}{HTML}{B28CFF}

\newcommand{\webagents}{\textcolor{domainred}{\textbf{Web Agents}}\xspace}
\newcommand{\egocentricvideos}{\textcolor{domaingreen}{\textbf{Egocentric Videos}}\xspace}
\newcommand{\roboticsdomain}{\textcolor{domainorange}{\textbf{Robotics}}\xspace}
\newcommand{\gamesdomain}{\textcolor{domainblue}{\textbf{Games}}\xspace}
\newcommand{\spatialunderstanding}{\textcolor{domainpurple}{\textbf{Spatial Planning}}\xspace}

\title{\centering \fontsize{18}{24}\selectfont{
\model{}: A Unified Benchmark for \\Agentic Visual Understanding
}}

\author{
    \vspace{.2cm}
    \parbox{\textwidth}{\centering
        \Authfont
        Advait Gupta\textsuperscript{1} \hspace{.15em}
        Sachin Konan\textsuperscript{2}
        \hspace{.15em}
        Tom Yin\textsuperscript{2} \hspace{.15em}
        Gabriel Sarch\textsuperscript{2*} \hspace{.15em}
        Zhuang Liu\textsuperscript{2*}
    }
    \\
    \vspace{.3cm}
    {\normalfont\fontsize{11}{15}\selectfont
    {\textsuperscript{1}University of Maryland, College Park}\hspace{.25cm}}
    {\normalfont\fontsize{11}{15}\selectfont
    {\textsuperscript{2}Princeton University}}\\
    \vspace{.3cm}
    $\vcenter{\hbox{\includegraphics[height=1.2em]{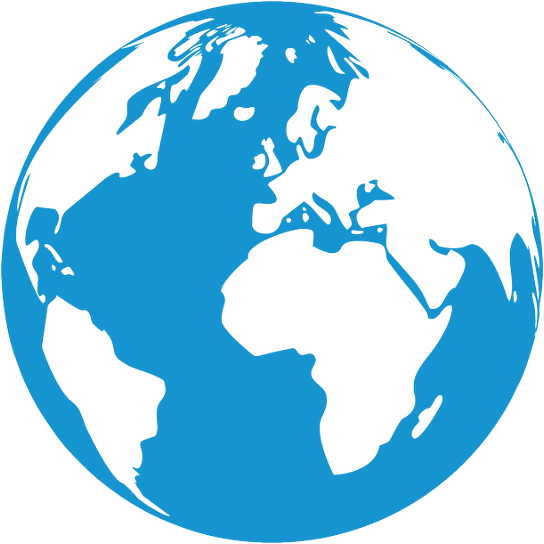}}}$\hspace{0.3em}
    \href{https://zlab-princeton.github.io/AgentVQA}{\texttt{Website}}
    \vspace{0.2cm} \\
    \raisebox{-0.4ex}{\includegraphics[height=1em]{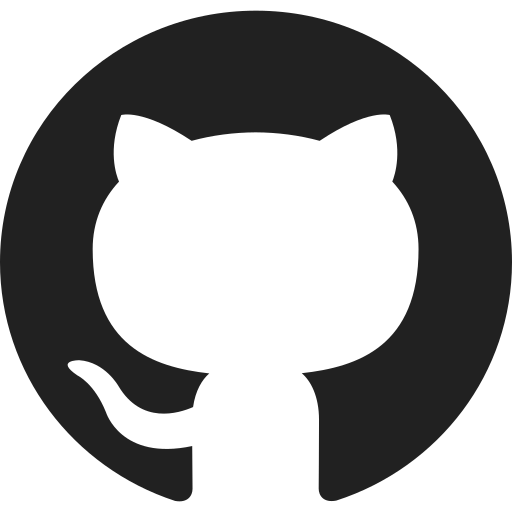}}\hspace{0.3em}
    \href{https://github.com/zlab-princeton/AgentVQA}{\texttt{Code}}
    \hspace{0.2cm}
    \raisebox{-0.4ex}{\includegraphics[height=1em]{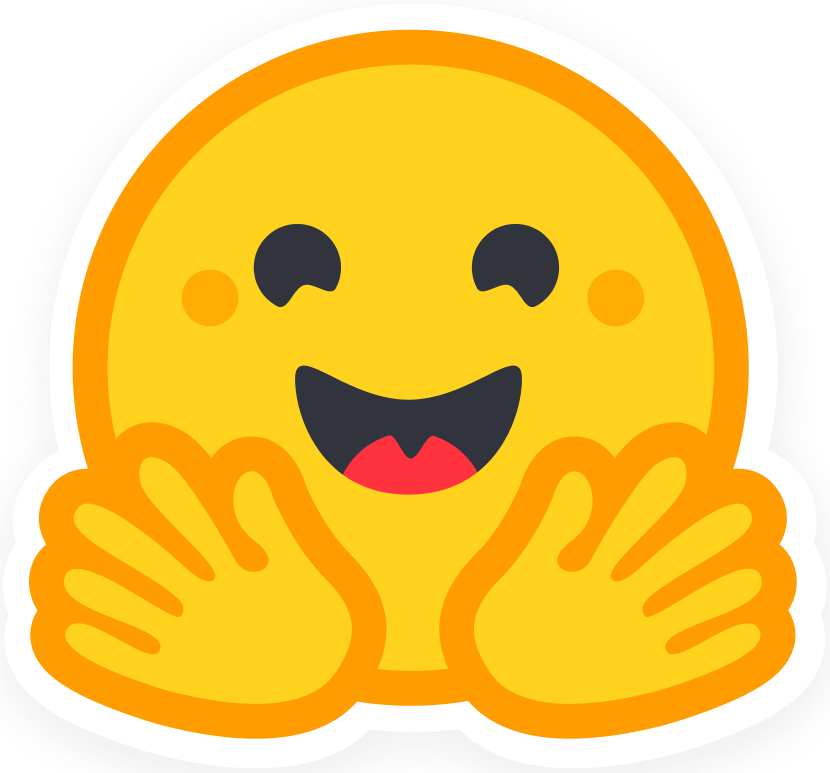}}\hspace{0.3em}
    \href{https://huggingface.co/datasets/AgentVQA/AgentVQA}{\texttt{Data}}
    \vspace{-3mm}
}

\makeatletter
\renewcommand{\abscontent}{}%
\makeatother

\newenvironment{abstractblock}{%
  {\centering\large\bfseries\sffamily Abstract\par}
  \vspace{0.2em}
  \begin{list}{}{%
      \setlength{\leftmargin}{2em}
      \setlength{\rightmargin}{2em}
      \setlength{\topsep}{0pt}
      \setlength{\parsep}{0pt}
  }
  \item[]
}{%
  \end{list}
  \par\normalfont\vspace{1em}
}

\makeatletter
\fancypagestyle{firststyle}{
  \fancyhead[L]{}
  \fancyhead[C]{}
  \fancyhead[R]{}
  \fancyfoot[L]{\footerfont \the\correspondingauthor}
  \fancyfoot[R]{}
}
\makeatother

\makeatletter
\DeclareRobustCommand\bfseries{%
  \not@math@alphabet\bfseries\mathbf
  \fontseries\bfdefault\selectfont
  \sffamily
}
\makeatother

\DeclareTextFontCommand{\textbf}{\bfseries\sffamily}
\providecommand{\teaserfilepath}{figures/F1_polished.pdf}

\newcommand{\teaserfigure}[2]{%
    \begin{figure}[#1]
        \centering
        \includegraphics[width=#2]{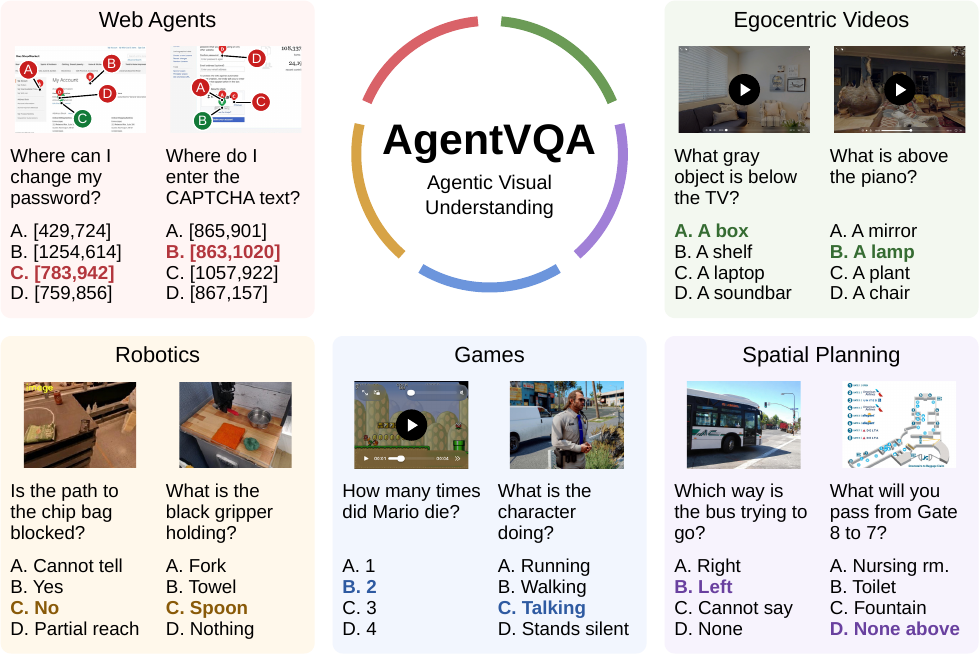}
        \caption{\textbf{The AgentVQA Benchmark.} AgentVQA unifies five challenging agentic domains: \webagents, \egocentricvideos, \roboticsdomain, \gamesdomain, and \spatialunderstanding.
        }
        \label{fig:teaser}
    \end{figure}%
}

\ifdefined\firstpagetrial\else
    \def\teaseraboveabstract{1}
\fi

\newcommand{\firstpagetitlegap}{4pt}
\newcommand{\firstpageauthorlinegap}{2pt}
\newcommand{\firstpagebeforelinksgap}{2pt}
\newcommand{\firstpageafterlinksgap}{3pt}
\newcommand{\firstpageabstractgap}{3pt}
\newcommand{\firstpagecaptionskip}{3pt}
\ifdefined\teaseraboveabstract
    \renewcommand{\firstpagetitlegap}{10pt}
    \renewcommand{\firstpageauthorlinegap}{4pt}
    \renewcommand{\firstpagebeforelinksgap}{16pt}
    \renewcommand{\firstpageafterlinksgap}{12pt}
    \renewcommand{\firstpageabstractgap}{5pt}
    \renewcommand{\firstpagecaptionskip}{6pt}
\fi

\author{%
    {\normalfont\fontsize{11}{12}\selectfont
    Advait Gupta\textsuperscript{1}\quad
    Sachin Konan\textsuperscript{2}\quad
    Tom Yin\textsuperscript{2}\quad
    Gabriel Sarch\textsuperscript{2*}\quad
    Zhuang Liu\textsuperscript{2*}\par
    \vspace{\firstpageauthorlinegap}
    \textsuperscript{1}University of Maryland, College Park\quad
    \textsuperscript{2}Princeton University\par
    \vspace{\firstpagebeforelinksgap}
    \raisebox{-0.2ex}{\includegraphics[height=1em]{figures/world.jpg}}\,
    \href{https://zlab-princeton.github.io/AgentVQA}{\texttt{Website}}\quad
    \raisebox{-0.2ex}{\includegraphics[height=1em]{figures/github.png}}\,
    \href{https://github.com/zlab-princeton/AgentVQA}{\texttt{Code}}\quad
    \raisebox{-0.2ex}{\includegraphics[height=1em]{figures/huggingface.png}}\,
    \href{https://huggingface.co/datasets/AgentVQA/AgentVQA}{\texttt{Data}}\par}
}

\makeatletter
\renewcommand{\maketitle}{%
    \begingroup
    \setlength{\parskip}{0pt}
    \centering
    {\titlefont\@title\par}
    \vspace{\firstpagetitlegap}
    {\@author\par}
    \endgroup
    \vspace{\firstpageafterlinksgap}
    \thispagestyle{firststyle}
}
\makeatother

\renewenvironment{abstractblock}{%
    \begingroup
    \setlength{\parskip}{0pt}
    {\centering\large\bfseries\sffamily Abstract\par}
    \vspace{\firstpageabstractgap}
    \noindent
}{%
    \par\endgroup\vspace{2pt}
}

\let\standardteaserfigure\teaserfigure
\renewcommand{\teaserfigure}[2]{%
    \begingroup
    \setlength{\intextsep}{2pt}
    \captionsetup{skip=\firstpagecaptionskip}
    \standardteaserfigure{#1}{#2}
    \endgroup
    \ifdefined\teaseraboveabstract
        \par
    \else
        \clearpage
    \fi
}

\begin{document}

\begingroup
\makeatletter
\let\raggedright\centering
\makeatother
\maketitle
\endgroup

\begingroup\renewcommand\thefootnote{}\footnotetext{\fontsize{9}{12}\selectfont$^*$Equal advising.}\addtocounter{footnote}{-1}\endgroup

\ifdefined\teaseraboveabstract
    \teaserfigure{H}{0.9\textwidth}
\fi

\begin{abstractblock}
Vision-language models (VLMs) can perform a broad range of tasks across diverse settings, yet their performance in agentic contexts remains poorly understood. Although online evaluations in simulators and real-world environments are essential for measuring interactive agent performance, they are costly and challenging to scale. Complementary offline VQA benchmarks can enable faster and more reliable comparisons across agentic domains, yet existing benchmarks remain limited and fragmented. To address this gap, we introduce AgentVQA, an offline benchmark for evaluating VLMs on realistic agentic scenarios across five agent domains: \webagents, \roboticsdomain, \egocentricvideos, \gamesdomain, and \spatialunderstanding. We build AgentVQA by manually selecting 1000 challenging, real-world agent examples from 23 source datasets and adapting them for reliable offline evaluation. The resulting benchmark preserves realistic scenarios from interactive environments across diverse domains, and its model rankings track those from corresponding online evaluations. Across the 22 evaluated models, the best achieves only 58.5\% accuracy. Our cross-domain analysis shows that errors are driven primarily by failures to ground and maintain accurate representations of the current and evolving visual state, including localizing relevant UI elements, interpreting motion, tracking state changes over time, and inferring 3D spatial relationships, rather than by task misinterpretation or downstream logical reasoning.


\end{abstractblock}

\ifdefined\teaseraboveabstract
    \clearpage
\else
    \teaserfigure{H}{\textwidth}
\fi

\section{Introduction}
Vision-Language Models (VLMs) are quickly becoming the decision-making core for agentic systems spanning robotics, web navigation, or gameplay. By prompting or fine-tuning, VLMs often outperform their domain-specific counterparts~\citep{koh2024visualwebarena, black2410pi0,zhou2025vision,zhang2025videogamebench,sarch2025grounding} and serve as the backbone for top-performing action models~\citep{pi0,pi0.5,ji2025robobrainunifiedbrainmodel}.

In domains like mathematics and coding, standardized benchmarks have accelerated progress by systematically tracking model capabilities and reliably exposing performance gaps~\citep{jimenez2023swe, balunovic_srimatharena_2025}. For agents, online evaluations in real or simulated environments remain the gold standard for assessing agentic performance \citep{yao2024taubenchbenchmarktoolagentuserinteraction}, but due to computational cost and reproducibility, they present practical challenges for rapid, large-scale model comparison~\citep{henderson2018deep,dasari2022rb2}.

As a result, VLMs' agentic capabilities lack comparable evaluation frameworks. State-of-the-art models are often assessed on general-purpose benchmarks like MMMU~\citep{yue2024mmmu}, MMBench~\citep{liu2024mmbench} and GPQA~\citep{rein2024gpqa}. While these benchmarks assess broad vision-language and language understanding, their questions and visual inputs do not reflect agentic decision-making scenarios. They primarily assess general visual understanding or knowledge-intensive reasoning. Recent works~\citep{yao2024taubenchbenchmarktoolagentuserinteraction, yehudai2025surveyevaluationllmbasedagents, wong2025widesearchbenchmarkingagenticbroad} show that performance on general-purpose VQA benchmarks does not reliably correlate with success in agentic tasks.

To address this, several domain-specific benchmarks have advanced evaluation of agentic capabilities within their respective settings~\citep{koh2024visualwebarena, rawles2024androidworld,bonatti2025windows,jang2025videowebarenaevaluatinglongcontext,pmlr-v267-yang25f}. However, these often provide narrow domain coverage, inputs and action spaces that do not reflect real agentic scenarios, or evaluation sets that are either saturated or affected by noisy annotations.

We introduce \textbf{\model{}}, a manually curated benchmark for systematically evaluating general-purpose VLMs on realistic agentic scenarios. \model{} contains 1,000 single-image, multi-frame, and video-based multiple-choice questions curated from 23 datasets across five domains: \webagents, \egocentricvideos, \roboticsdomain, \gamesdomain, and \spatialunderstanding. These domains represent distinct forms of agent--environment interaction, including GUI navigation, temporally extended first-person understanding, embodied perception and action, strategic decision-making, and 3D spatial planning, as illustrated in Figure~\ref{fig:teaser}. By evaluating these settings within a common benchmark, \model{} supports both aggregate model comparison and analysis of whether agentic capabilities generalize across domains.

We construct \model{} through a human-centered, model-assisted pipeline designed to provide broad coverage, standardized evaluation, and meaningful difficulty. VLMs screen candidates and propose hard distractors, while annotators select, repair, and verify every retained example. Only about 22\% of the initial candidates survive the full curation pipeline, as human review identifies subtle ambiguities and annotation errors missed by automated checks and removes examples that are insufficiently challenging. The final benchmark expresses heterogeneous targets, including coordinates, actions, spatial relations, and predicted states, through a common multiple choice interface while preserving the semantics of the original tasks.

Model rankings on \model{} correlate with those from SpatialWorld~\citep{gao2026spatialworldbenchmarkinginteractivespatial}, an online evaluation of agentic capabilities in simulation, supporting its use as a scalable complement to interactive evaluation. At the same time, its rankings differ substantially from those on broad visual and reasoning benchmarks such as MMMU-Pro, GPQA-Diamond, and ChartQAPro, suggesting that performance on conventional benchmarks does not reliably predict agentic competence. Across the 22 open- and closed-source models we evaluate, the strongest model achieves only 58.5\% accuracy. Moreover, the strongest model differs across domains, showing that aggregate performance can obscure substantial variation in domain-specific capabilities.

Our cross-domain error analysis reveals a shared limitation that manifests differently across agentic settings. Failures are dominated by grounded scene and state understanding: models struggle with UI localization in \webagents, motion interpretation in \egocentricvideos, 3D spatial reasoning in \roboticsdomain and \spatialunderstanding, and visual state comprehension in \gamesdomain. In contrast, failures arising from misunderstanding the task or selecting an answer inconsistent with the model's reasoning are rare. These results suggest that current VLMs often understand the requested objective and can reliably reason over perceived inputs but fail to construct and maintain the grounded representation of the environment needed to act correctly.

\begin{figure}[h]

    \centering
        \includegraphics[width=\textwidth]{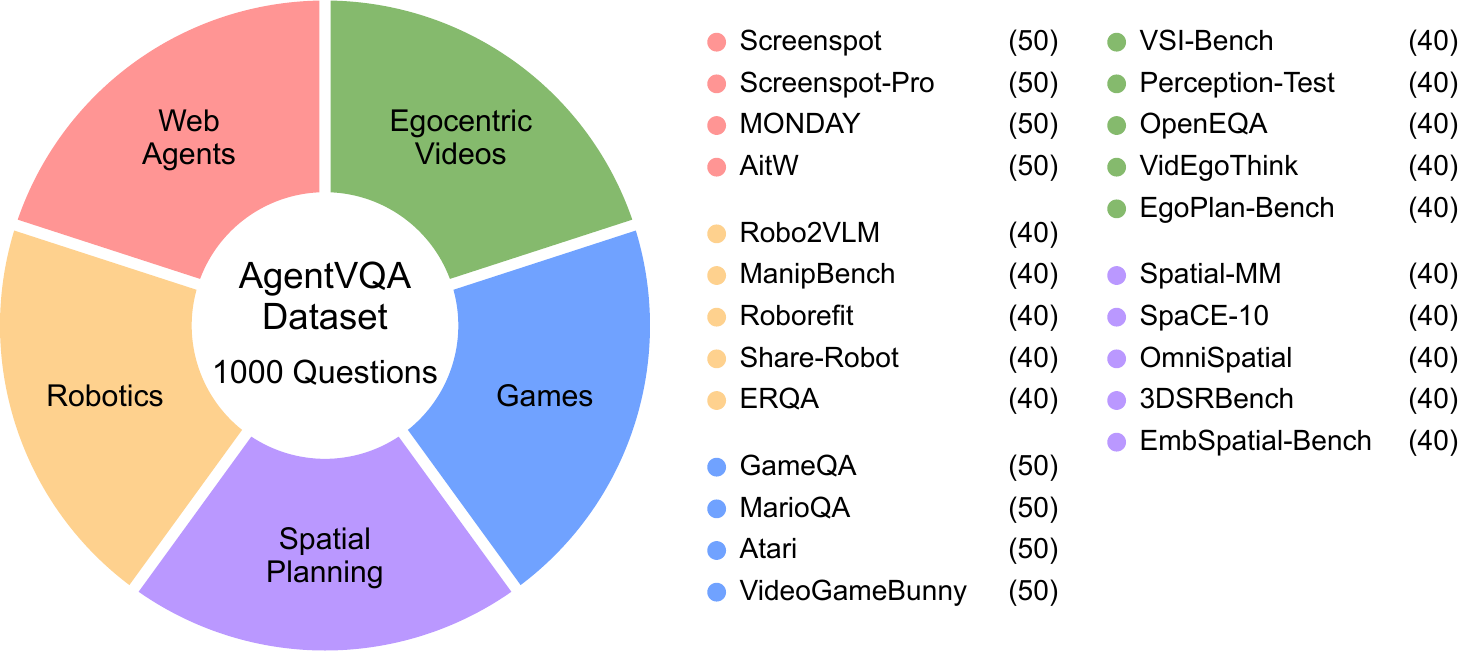}

    \caption{\textbf{Distribution of the 1000 questions in the AgentVQA benchmark.} Distribution across the 23 source datasets (sample counts in parentheses), organized within the 5 core agentic domains.}
    \label{fig:distribution} 
\end{figure}

\section{Related Work}
\textbf{VLM as agents.} Recent advances in multimodal AI have been driven by the development of Vision-Language Models (VLMs). These models, such as QwenVL~\citep{bai2023qwenvlversatilevisionlanguagemodel}, GPT-4o~\citep{openai2024gpt4ocard}, and Gemini~\citep{comanici2025gemini_base}, are typically pretrained on vast, web-scale datasets of paired images and text. This equips them with a broad general-purpose understanding of both visual concepts and natural language, making them an ideal compute unit for agents. Several works have successfully adapted VLMs into specific domains, either by fine-tuning or scaffolding pre-trained models~\citep{black2410pi0,koh2024visualwebarena,zhou2025vision,zhang2025videogamebench,sarch2025grounding}. AgentVQA evaluates agentic capabilities across domains.

\textbf{General visual question answering.} A wealth of benchmarks have been developed to assess the general capabilities of VLMs, including SEED-Bench~\citep{li2023seed}, MMBench~\citep{liu2024mmbench}, or MMMU~\citep{yue2024mmmu}. While invaluable for measuring core competencies and reasoning, they are ill-suited as proxies for agentic intelligence. Agentic tasks often require episodic memory~\citep{majumdar2024openeqa} because either the modalities are iterative, e.g., a stream of image observations, or the task requires interactive reasoning, e.g., tool-calling.~\citet{yao2024taubenchbenchmarktoolagentuserinteraction} found that GPT-4o, despite excelling on traditional benchmarks, achieves a success rate less than 50\% on realistic customer service tasks requiring multi-turn interactions. In Fig.~\ref{fig:rank_correlation}, we confirm this observation by the divergence between rankings in AgentVQA vs MMMU-Pro and GPQA-Diamond.

\textbf{Agentic benchmarks.} The agentic evaluation landscape spans language-only benchmarks, e.g., SWE-bench~\citep{jimenez2023swe} and AgentBench~\citep{liu2023agentbenchevaluatingllmsagents}, which test capabilities such as code generation and tool use, and vision-language benchmarks requiring visual grounding and spatial reasoning. These benchmarks can also be categorized as online or offline. Online benchmarks, e.g., \citep{savva2019habitatplatformembodiedai, fan2022minedojobuildingopenendedembodied, li2024behavior1khumancenteredembodiedai, koh2024visualwebarena, zhang2025videogamebench}, evaluate agents through environment interaction and therefore capture non-deterministic dynamics important for real-world deployment, but are often costly and infrastructure-intensive~\citep{henderson2018deep, du2024embspatialbenchbenchmarkingspatialunderstanding}. Offline benchmarks, e.g., \citep{rawles2023androidwildlargescaledataset, yang2025thinking, comanici2025geminirobotics, tong2025gamerlsynthesizingverifiablegame, gong2025space10comprehensivebenchmarkmultimodal}, instead use verifiable question-answer tasks paired with visual inputs. However, existing offline benchmarks are largely domain-specific and differ in answer formats, option difficulty, and sample distributions, making direct aggregation and cross-domain analysis difficult (Table~\ref{tab:agentic_benchmarks_compact}). AgentVQA is curated by filtering and transforming samples into a standardized, unambiguous multiple choice format with hard negatives.

\begin{figure}[t]
    \centering
    \includegraphics[width=\textwidth]{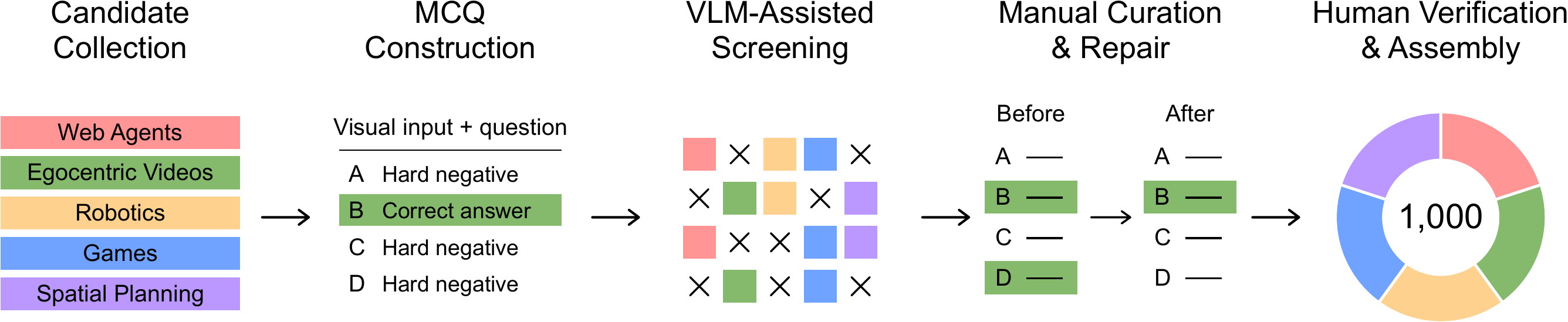}
\caption{\textbf{Dataset curation pipeline for AgentVQA.} (1) Collect candidates across 23 datasets, task types, and input modalities spanning five agentic domains. (2) Standardize tasks as MCQs; Gemini~2.5 Pro proposes three hard negatives when needed. (3) Gemini and Claude flag ambiguity, annotation errors, and weak distractors, removing approximately 45\% of candidates. (4) Human annotators review, edit, and repair the remaining samples, rejecting approximately 60\% as ambiguous, incorrect, or too easy. (5) Verify exactly one correct answer and plausible, unambiguously incorrect hard negatives, then assemble 1,000 manually verified questions balanced across the five domains.}\label{fig:curation_pipeline}
\end{figure}

\begin{figure}[t]
    \centering
    \includegraphics[width=0.9\textwidth]{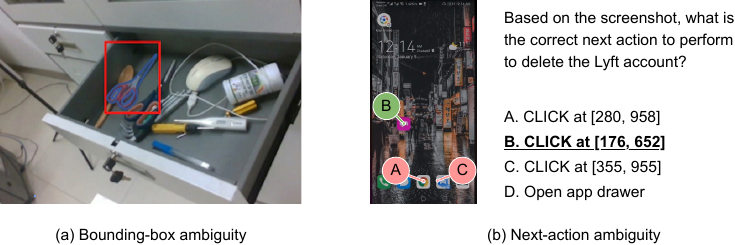}
    \caption{\textbf{Ambiguities addressed by manual curation (Stage~4, Section~\ref{sec:construction}).} (a) A rectangular bounding box around scissors includes background, allowing clicks that miss the object to be scored as correct. (b) For the goal of deleting a Lyft account, B (green) directly opens Lyft and is the intended next action. However, D can also lead to a valid solution by opening the app drawer and then launching Lyft, adding a navigation step. The question does not require the shortest route, leaving multiple plausible next actions. Manual review removes or repairs such ambiguous samples to ensure a unique correct answer.}
    \label{fig:bbox_failures}
\end{figure}

\section{The AgentVQA Benchmark}
\label{sec:AgentVQA}


We define agentic tasks as those that evaluate action selection or the perceptual and reasoning capabilities required for goal-directed interaction with an environment~\cite{bengio2025superintelligentagentsposecatastrophic}. To support scalable offline evaluation of these capabilities, AgentVQA comprises 1,000 manually curated multiple-choice questions drawn from 23 datasets spanning five agentic domains. Its construction is guided by three principles: \textbf{comprehensive} coverage across domains to assess skill generality; \textbf{standardized} evaluation using manually curated questions with unambiguous ground truth; and \textbf{challenging} examples, states, and negative options designed to expose fine-grained weaknesses in current models.

To provide comprehensive coverage, AgentVQA spans five distinct forms of agent-environment interaction: digital interaction and GUI navigation in \webagents; embodied perception and action in \roboticsdomain; temporally extended first-person planning in \egocentricvideos; strategic decision-making in \gamesdomain; and spatial navigation and planning in \spatialunderstanding. The benchmark is balanced across domains, with 200 questions per domain (Figure~\ref{fig:distribution}). Its inputs range from individual screenshots and embodied-scene images to multi-frame trajectories, egocentric videos, and gameplay videos. Heterogeneous targets, including coordinates, actions, spatial relations, and predicted states, are represented through a common multiple-choice format while preserving observations of the agentic scenario. See Appendix~\ref{sec:appendix_dataset_overview} for full details of the datasets.

\vspace{-1em}
\subsection{Data Curation and Construction}\label{sec:construction}
\vspace{-0.5em}

We construct AgentVQA through a five-stage, human-centered pipeline. VLMs assist with large-scale screening and hard-negative generation, while human annotators determine which examples are retained and verify the correctness, clarity, and difficulty of every final question.

\textbf{Stage 1: Candidate collection.} We assemble a diverse candidate pool from 23 source datasets spanning five agentic domains (see Appendix~\ref{sec:appendix_dataset_overview} for full dataset list). Candidates are sampled across datasets, task types, and input modalities to cover a broad range of realistic observations, states, and decisions, while initially retaining their original questions, annotations, and answer formats.

\textbf{Stage 2: MCQ construction.}
\label{sec:mcq_conversion}
We convert the candidate tasks into a common multiple-choice format. For examples that are not originally multiple choice, Gemini 2.5 Pro proposes three hard negative options using the visual input, task prompt, correct answer, action history, and relevant task metadata. The generated distractors include both near-miss negatives, which differ from the correct answer in a fine-grained way, and semantically similar negatives, which correspond to plausible but incorrect objects, actions, or locations.

\textbf{Stage 3: VLM-assisted screening.} We use a multi-turn verification pipeline with Gemini and Claude to identify converted questions with ambiguous instructions, inconsistent annotations, multiple plausible answers, malformed states, implausible distractors, or other quality issues. This high-recall screening removes approximately 45\% of the initial candidate pool, reducing the amount of manual review required while preserving human judgment for cases involving subtle visual or domain-specific ambiguities.

\textbf{Stage 4: Manual curation and repair.}
\label{sec:manual_curation}
Human annotators then inspect the remaining candidates for correctness, difficulty, diversity, and relevance to agentic decision-making. Approximately 60\% of the samples that pass automated screening are still rejected because they contain subtle ambiguities, annotation errors, invalid distractors, or are insufficiently challenging. When possible, annotators repair promising examples by refining the question, ground-truth answer, visual input, answer choices, or associated metadata before retaining them for the final evaluation format.

\textbf{Stage 5: Human verification and benchmark assembly.} Every completed MCQ is manually reviewed a second time to ensure that exactly one answer is correct and that the distractors are plausible, challenging, and unambiguously incorrect. This step along with Manual Curation is especially important for grounding tasks, where bounding boxes and automatic spatial checks can admit false positives or reject valid interaction points, as illustrated in Figure~\ref{fig:bbox_failures}. We revise questions, annotations, and options as needed, then assemble a balanced final benchmark of 1,000 manually verified questions, with 200 examples per domain and 40 or 50 examples from each source dataset.

\section{Experiments}

In this section, we investigate the ability of different models to perform agentic visual tasks across the diverse inputs curated in AgentVQA. We first describe the evaluation setup and model suite in Section~\ref{sec:experimental_setup}. We then report overall model performance across domains, domain-level trends, action-history effects, open- versus closed-source models, and reasoning variants in Section~\ref{sec:overall_performance}, compare AgentVQA rankings with existing offline benchmarks in Section~\ref{sec:rank_correlation}, and assess its reliability as a lower-cost alternative to online agent evaluation through a comparison with SpatialWorld in Section~\ref{sec:online_correlation}. Finally, we examine the failure modes in Section~\ref{sec:error_modes}.


\subsection{Experimental Setup}
\label{sec:experimental_setup}
\textbf{Models.} We evaluate a diverse suite of 22 prominent VLMs to ensure a comprehensive assessment of the current landscape. Our selection includes closed-source models: GPT-5.6-sol thinking-high/none~\citep{openai2025gpt5}, GPT-5.6-luna thinking-high/none~\citep{openai2025gpt5}, GPT-4o~\citep{openai2024gpt4ocard}, Gemini 3.1 Pro~\citep{gemini3} and 3.7 Flash~\citep{comanici2025gemini_base}, Claude Opus 4.8~\citep{opus} and Sonnet 5~\citep{sonnet}, and Qwen3.7 Plus and Flash~\citep{qwen37}. We also include a wide range of open-source models: the GLM series (4.6V~\citep{glm46}, 4.5V~\citep{vteam2026glm45vglm41vthinkingversatilemultimodal}, 4.1V-9B-Thinking/Base~\citep{vteam2026glm45vglm41vthinkingversatilemultimodal}), Kimi K3~\citep{kimi}, Llama 4 Maverick~\citep{meta2024llama4}, the Qwen3.5 series (397B-A17B, 122B-A10B, 35B-A3B, 27B)~\citep{qwen35}, and Qwen3.8 27B~\citep{qwen38}.

Models are categorized as ``Reasoning'' or ``Non-Reasoning'' based on whether they allocate extended token budgets for internal chain-of-thought reasoning before producing a final answer.

\textbf{Evaluation setup.} For models that do not natively support video input, we sample 32 frames at evenly spaced intervals. This number is validated by our own ablation studies and prior work~\citep{yang2025thinking}. Models are evaluated with a temperature of 0.8 and 0.2 for reasoning and non-reasoning models, respectively. For models without manual temperature like GPT series, we use the default.

We use a standard prompt that provides the question and any relevant context, querying the model to output only the single letter corresponding to its choice. For questions involving spatial coordinates, options are formatted in each model's native convention (e.g., normalized [0, 1000] in (y, x) order for Gemini, absolute pixel coordinates for Qwen). Final answers are extracted programmatically, and all visual inputs are provided at native resolution. Further details on prompts and answer extraction are in Appendices~\ref{sec:appendix_eval_prompt} and~\ref{sec:appendix_extraction_logic}.

\newcolumntype{C}[1]{>{\centering\arraybackslash}p{#1}}

\begin{table}[h!]
\centering
\caption{\textbf{Performance of VLMs on AgentVQA across the five agentic domains.}\\ \scalebox{1.4}{\twemoji{thinking face}} represents Reasoning Models. \scalebox{1.4}{\twemoji{locked with key}} represents closed-source models. Bolded and underlined numbers indicate best and second-best rank per column, respectively.\\}
\label{tab:my_label_custom_width}
\setlength{\tabcolsep}{2pt}
\footnotesize
\makebox[\textwidth][c]{%
\begin{tabular}{@{}
    p{3.7cm} 
    C{1.25cm} 
    C{1.5cm} 
    C{1.25cm} 
    C{1.25cm} 
    C{1.25cm} 
    C{1.25  cm} 
@{}}

\toprule
\textbf{Model} & \textbf{\textcolor{domainred}{Web}} & \textbf{\textcolor{domaingreen}{Egocentric}} & \textbf{\gamesdomain} & \textbf{\textcolor{domainpurple}{Spatial}} & \textbf{\roboticsdomain} & \textbf{Avg} \\
\textbf{} & \textbf{\textcolor{domainred}{Agents}} & \textbf{\textcolor{domaingreen}{Videos}} & \textbf{} & \textbf{\textcolor{domainpurple}{Planning}} & \textbf{} & \textbf{} \\
\midrule
GPT 5.6-sol think-high \scalebox{0.8}{\twemoji{thinking face}} \scalebox{0.8}{\twemoji{locked with key}} & \textbf{47.0} & \underline{69.0} & 62.5 & 51.0 & 45.5 & 55.0 \\
GPT 5.6-sol think-none \scalebox{0.8}{\twemoji{locked with key}} & 16.0 & 59.0 & 44.5 & 41.0 & 30.5 & 38.2 \\
GPT 5.6-luna think-high \scalebox{0.8}{\twemoji{thinking face}} \scalebox{0.8}{\twemoji{locked with key}} & 34.5 & 65.0 & 61.0 & 43.5 & 35.5 & 47.9 \\
GPT 5.6-luna think-none \scalebox{0.8}{\twemoji{locked with key}} & 19.5 & 46.0 & 37.5 & 32.0 & 29.0 & 32.8 \\
GPT-4o \scalebox{0.8}{\twemoji{locked with key}} & 21.5 & 40.0 & 29.0 & 35.0 & 31.5 & 31.4 \\
\midrule
Gemini 3.1-pro \scalebox{0.8}{\twemoji{thinking face}} \scalebox{0.8}{\twemoji{locked with key}} & 34.5 & 64.0 & \underline{68.0} & \underline{61.0} & \underline{51.0} & \underline{55.7} \\
Gemini 3.7-flash \scalebox{0.8}{\twemoji{locked with key}} & 40.5 & 67.0 & \textbf{70.0} & \textbf{62.5} & \textbf{52.5} & \textbf{58.5} \\
\midrule
Claude Opus 4.8 \scalebox{0.8}{\twemoji{locked with key}} & 35.5 & 60.0 & 50.5 & 41.0 & 33.0 & 44.0 \\
Claude Sonnet 5 \scalebox{0.8}{\twemoji{locked with key}} & 38.0 & 47.0 & 53.0 & 42.5 & 36.5 & 43.4 \\
\midrule
GLM 4.6v \scalebox{0.8}{\twemoji{thinking face}} & 25.5 & 54.0 & 47.5 & 39.5 & 42.5 & 41.8 \\
GLM 4.5v \scalebox{0.8}{\twemoji{thinking face}} & 25.0 & 65.0 & 46.5 & 41.0 & 37.0 & 42.9 \\
GLM 4.1V 9B-Thinking \scalebox{0.8}{\twemoji{thinking face}} & 19.0 & 53.0 & 41.0 & 38.5 & 25.5 & 35.4 \\
GLM 4.1V 9B-Base & 20.5 & 46.0 & 37.5 & 44.5 & 33.0 & 36.3 \\
\midrule
Kimi K3 \scalebox{0.8}{\twemoji{thinking face}} & 22.0 & 64.0 & 63.5 & 53.0 & 39.0 & 48.3 \\
\midrule
Llama-4-Maverick & 18.5 & 49.0 & 23.0 & 30.0 & 19.5 & 28.0 \\
\midrule
Qwen3.8 27B \scalebox{0.8}{\twemoji{thinking face}} & 30.0 & \underline{69.0} & 61.5 & 51.5 & 42.5 & 50.9 \\
Qwen3.7 Plus \scalebox{0.8}{\twemoji{thinking face}}\scalebox{0.8}{\twemoji{locked with key}} & 32.5 & 60.0 & 61.5 & 50.0 & 39.0 & 48.6 \\
Qwen3.7 Flash \scalebox{0.8}{\twemoji{thinking face}}\scalebox{0.8}{\twemoji{locked with key}} & \underline{43.0} & \textbf{72.5} & 61.0 & 55.0 & 43.5 & 55.0 \\
Qwen3.5 397B-A17B \scalebox{0.8}{\twemoji{thinking face}} & 25.0 & 65.0 & 57.5 & 51.5 & 40.0 & 47.8 \\
Qwen3.5 122B-a10b \scalebox{0.8}{\twemoji{thinking face}} & 26.0 & 60.0 & 53.0 & 50.0 & 41.0 & 46.0 \\
Qwen3.5 35B-a3b \scalebox{0.8}{\twemoji{thinking face}} & 34.5 & 62.0 & 55.5 & 45.5 & 42.0 & 47.9 \\
Qwen3.5 27B \scalebox{0.8}{\twemoji{thinking face}} & 32.5 & 57.0 & 49.5 & 51.5 & 43.0 & 46.7 \\
\bottomrule
\end{tabular}
}

\end{table}

\subsection{Results on AgentVQA}
\label{sec:overall_performance}

We report the results on AgentVQA in Table \ref{tab:my_label_custom_width}. We draw the following conclusions:

\textbf{Top models struggle with AgentVQA.}
Even the top performing model, Gemini 3.7 Flash, achieves only 58.5\% accuracy. A clear tiering is observable: the largest proprietary models (GPT-5.6, Gemini 3.7 Flash, Gemini 3.1 Pro) are at the top, followed by other competitive models such as Qwen3.7 Flash and the open-weight Qwen3.8 27B, with smaller and older-generation models in the lower tiers. Among the five domains, models perform best on \egocentricvideos and \gamesdomain, followed by \spatialunderstanding, \roboticsdomain, and worst on \webagents. This indicates that current VLMs are stronger at temporally contextualized visual understanding and game-state reasoning than at precise GUI grounding and spatial reasoning.

\begin{wrapfigure}{R}{0.4\columnwidth}
\centering
\includegraphics[width=\linewidth]{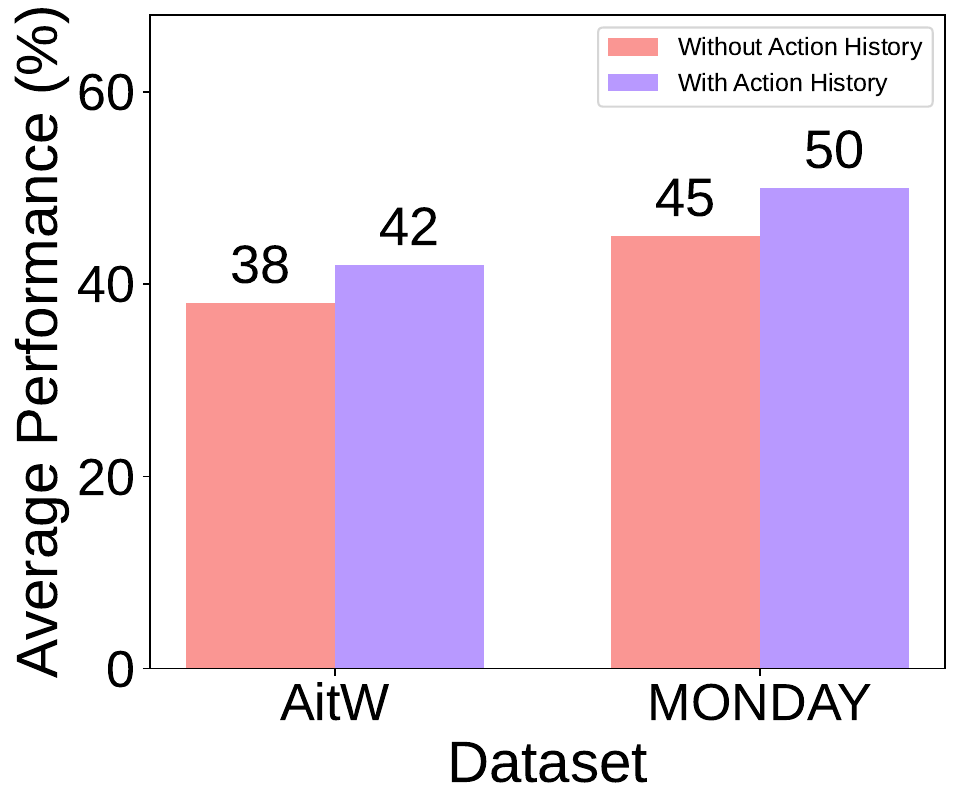}
\vspace{-2em}
\caption{Impact of action history on model performance in \webagents.}
\vspace{-2em}
\label{fig:action_history}
\end{wrapfigure}

\textbf{Large variance per domain.} Current VLMs are strongest on temporally grounded understanding (\egocentricvideos: 59\%, \gamesdomain: 51.5\%) and weaker on precise grounding and embodied reasoning (\spatialunderstanding: 46\%, \roboticsdomain: 38\%, \webagents: 29\%). \webagents is the hardest domain, with only three models exceeding 40\%. \gamesdomain exhibits the largest spread (23\% to 70\%), suggesting strategic reasoning is strong only in the most capable models. No single model dominates across all domains: Gemini 3.7 Flash leads \gamesdomain (70\%) but falls behind GPT-5.6-sol thinking-high in \webagents (40.5\% vs. 47\%), while Qwen3.7 Flash leads \egocentricvideos (72.5\%) and Gemini 3.7 Flash leads \spatialunderstanding (62.5\%) and \roboticsdomain (52.5\%). These cross-domain rank shifts indicate highly uneven capabilities across different settings.

\paragraph{Open-source models are catching up}
The gap between closed-source and open-source models remains narrow. The top closed-source model, Gemini 3.7 Flash (58.5\%), leads the top open-source model, Qwen3.8 27B (50.9\%), by 7.6\%. Kimi K3 (48.3\%), Qwen3.5 35B (47.9\%), and Qwen3.5 397B (47.8\%) outperform Claude Opus 4.8 (44.0\%) and Sonnet 5 (43.4\%). Multiple open-source models also substantially outperform GPT-5.6-sol thinking-none (38.2\%) and GPT-4o (31.4\%). These results suggest that while the strongest proprietary models retain an advantage on agentic tasks, competitive open-source alternatives already match many closed-source systems.


\paragraph{Reasoning offers varying results}
Extended reasoning helps some models but not others. GPT-5.6-sol gains 16.8 points with thinking-high, while GLM 4.1V performs worse with thinking as compared to without it. We find that GLM-4.1V-Thinking gets stuck in ``thinking loops'' (App. \ref{app:loops}).

\paragraph{Trajectory questions benefit from action history}
Our ablation on \webagents tasks reveals that providing action history is a consistent benefit (Figure \ref{fig:action_history}). Across five models (Gemini 3.1-pro, GPT-4o, Qwen3.5 27B, Kimi K3, and Claude Sonnet 5), removing the history decreases average performance by 9.5\% on AitW and 10\% on MONDAY. This confirms that access to historical context is critical for multi-step tasks.


\begin{figure}[h]
    \centering
    \includegraphics[width=1\textwidth]{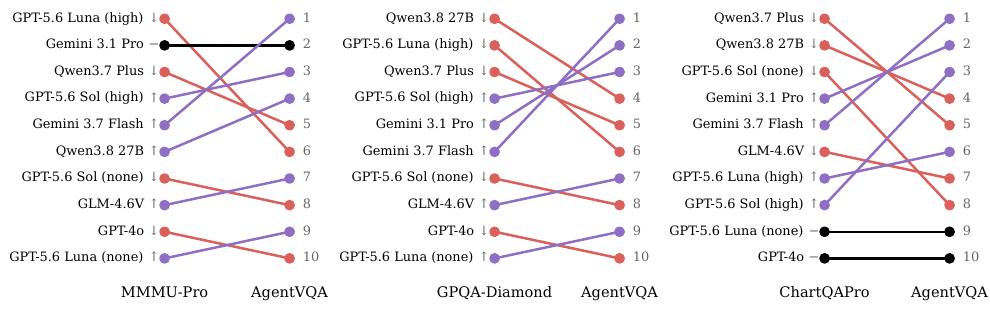}
    \caption{\textbf{Divergence in model performance rankings between AgentVQA and three existing benchmarks.} Lines are colored \textcolor{red}{red} if a model's rank degrades and \textcolor[HTML]{B28CFF}{purple} if it improves. The rank divergence shows that our benchmark has moderate-to-weak correlation with general VQA benchmarks. While a few models maintain their ranks, the relative positions of most others change. In the comparison with GPQA-Diamond, no model maintains its relative ranking.}
    
    \label{fig:rank_correlation}
    \end{figure}
\vspace{-2em}
\subsection{Rank Correlation}
\label{sec:rank_correlation}




\paragraph{Low correlation with existing benchmarks} To validate the benefit of a specialized agentic benchmark, we compare model rankings for a subset of models on AgentVQA to those on a subset of MMMU-Pro~\citep{yue2024mmmu} (250 samples), ChartQA-Pro~\citep{masry2025chartqaprodiversechallengingbenchmark} (250 samples), and GPQA-Diamond~\citep{rein2024gpqa} (150 samples), running our own evaluations with standardized prompts. As shown in Figure \ref{fig:rank_correlation}, rankings diverge substantially. Spearman's rank correlation coefficients are moderate for MMMU-Pro ($\rho = 0.67$) and GPQA-Diamond ($\rho = 0.59$), and weak for ChartQA-Pro ($\rho = 0.44$).

    


No single model dominates across all comparisons. Gemini 3.1 Pro ranks second on AgentVQA and MMMU-Pro but drops to 5th on GPQA-Diamond, while Qwen3.8 27B leads GPQA-Diamond but falls to 6th on MMMU-Pro. These shifts confirm that strong performance on general-purpose visual reasoning does not reliably predict agentic competence.

\begin{figure}[h]
    \centering
    \includegraphics[width=0.9\linewidth]{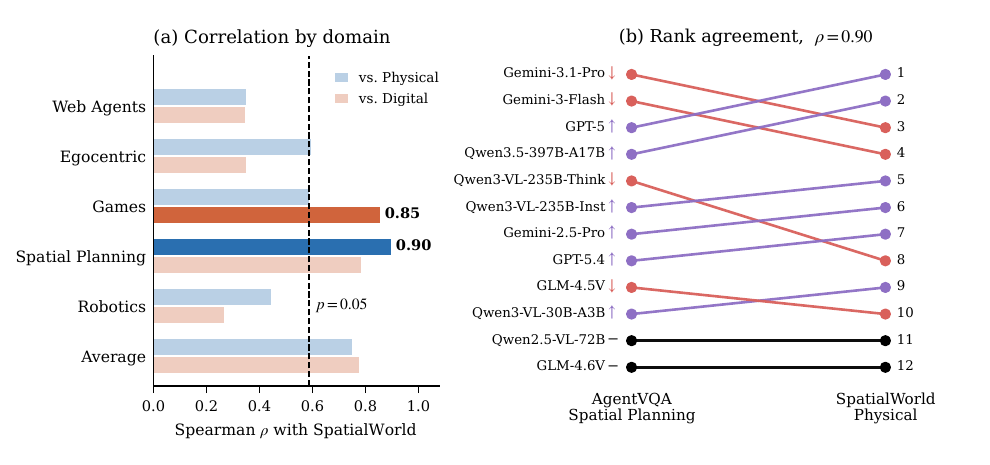}
    \caption{\textcolor{black}{\textbf{AgentVQA ranks models the way an online interactive benchmark does.} (a) Spearman correlation between each AgentVQA domain and the two SpatialWorld tracks over 12 shared models; saturated bars are the domain-matched pairs. (b) Model rankings under AgentVQA Spatial Planning against the SpatialWorld Physical track ($\rho = 0.895$, $p = 8\times10^{-5}$). \textcolor[HTML]{8E6FC4}{\textbf{Purple}} marks models that rank higher on SpatialWorld, \textcolor[HTML]{D9605B}{\textbf{red}} for models lower on it, and black unchanged.}}
    \label{fig:correlation_spatial}
\end{figure}
\subsection{Correlation with Online Agent Evaluations}
\label{sec:online_correlation}



To examine whether AgentVQA provides a reliable alternative to costly interactive evaluation, we compare its model rankings with performance on SpatialWorld~\citep{gao2026spatialworldbenchmarkinginteractivespatial}, a closed-loop benchmark for embodied and interactive agents. SpatialWorld scores are obtained from the official released evaluation results, which are produced under a single simulator-agnostic protocol across eight simulation backends. For this comparison, we separately evaluate 12 models from SpatialWorld on AgentVQA and compare their relative rankings via Spearman’s rank correlation. This analysis measures whether AgentVQA preserves the model ordering observed under interactive evaluation despite operating entirely offline and avoiding environment configuration and execution-related computational costs.

Figure~\ref{fig:correlation_spatial} shows a strong positive Spearman rank correlation of $\rho = \textbf{0.895}$ across \textbf{12} overlapping models between SpatialWorld and AgentVQA Spatial Planning, the closest category. This agreement demonstrates that AgentVQA provides a reliable and substantially lower-cost offline alternative for ranking agentic models, avoiding the environment setup, repeated rollouts, and computational expense required by online evaluation. The lower correlations for other AgentVQA domains suggest that model rankings depend on the capabilities being assessed, highlighting the limitations of a single dataset as a proxy for overall agentic performance. Together with the strong agreement in the matched domain, this pattern supports the interpretation that AgentVQA captures meaningful domain-specific performance differences.

\subsection{Error Modes}
\label{sec:error_modes}

To understand why models fail, we analyze incorrect predictions from Gemini~3.1~Pro, sampled evenly across datasets. We prompt the model to provide its reasoning with each answer, manually establish a taxonomy of common failure modes, and use Gemini to categorize the errors. We then manually review a randomly selected subset with 100 error mode assignments in total across the five domains.

\begin{figure}[h!]
    \centering
    \includegraphics[width=1\textwidth]{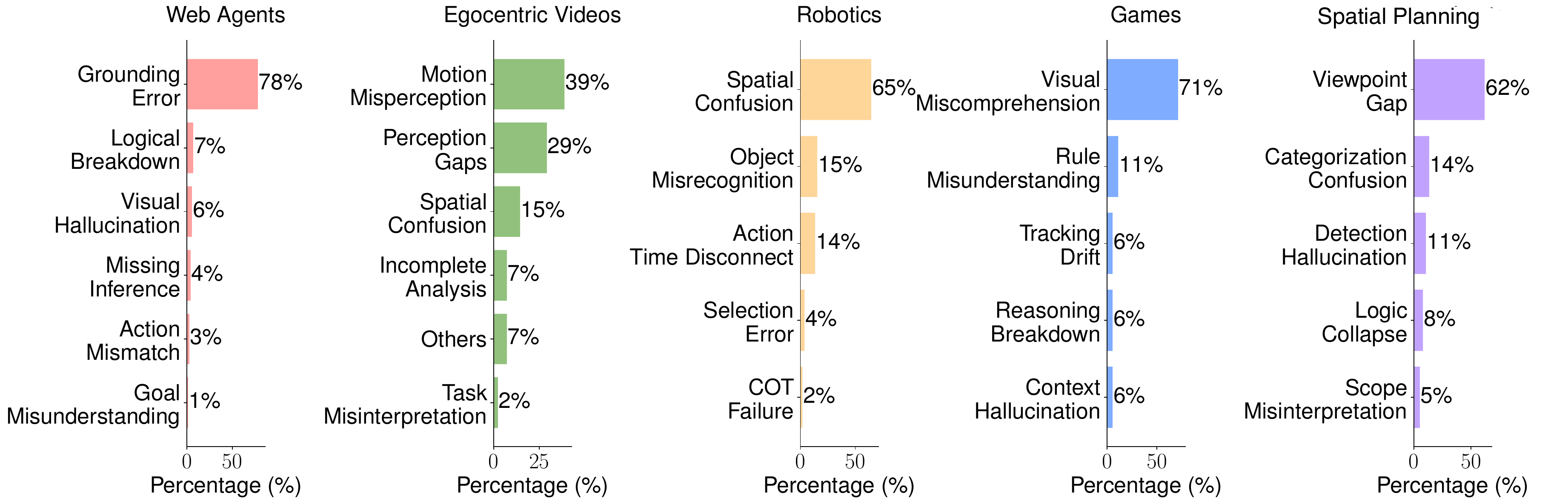}
    \caption{\textbf{Distribution of the most common error modes across the five agentic domains in AgentVQA.} The analysis is based on a semi-automated categorization of the incorrect predictions from Gemini-3.1-Pro. The chart is organized by domain, with each colored segment representing a primary domain. Each bar within a segment corresponds to a specific error mode, and its size reflects its prevalence (\%).}
    \label{fig:error_modes}
\end{figure}


For Gemini 3.1 Pro, the results reveal domain-specific manifestations of a shared weakness: constructing an accurate task-relevant state from visual observations (Figure~\ref{fig:error_modes}). Among the analyzed incorrect \webagents predictions, 78\% were classified as grounding errors, indicating difficulty locating the correct interface element. All percentages refer to the distribution of error categories among analyzed incorrect predictions within each domain. In \gamesdomain, visual miscomprehension (71\%) shows that failures primarily arise from reading the game state rather than applying its rules. \roboticsdomain and \spatialunderstanding are dominated by spatial confusion (65\%) and viewpoint gaps (62\%), respectively, exposing weaknesses in relational and perspective-aware reasoning. Errors in \egocentricvideos are more distributed across motion misperception (39\%), perception gaps (29\%), and spatial confusion (15\%), reflecting difficulty tracking dynamic states. Across domains, failures therefore arise mainly before high-level planning, during perception, grounding, and state construction. Additional details are provided in Appendix~\ref{sec:appendix_error_mode_methodology}.


\section{Conclusion}
We introduce AgentVQA, a unified offline benchmark that consolidates 23 datasets across five agentic domains into 1,000 manually verified multiple-choice questions with challenging hard negatives. While prior work typically evaluates agents within individual environments or domains~\citep{koh2024visualwebarena,rawles2024androidworld,bonatti2025windows,pmlr-v267-yang25f}, AgentVQA enables controlled comparison of visual decision-making capabilities across diverse agentic settings.

Our results show that these capabilities remain far from solved: among 22 state-of-the-art VLMs, the strongest model achieves only 58.5\% accuracy. Our analysis of Gemini 3.1 Pro’s incorrect predictions identifies grounding, perception, motion understanding, and spatial state construction as the predominant error categories. Action history improves performance, while reasoning benefits some models but not others. However, neither closes the remaining gap. Moreover, AgentVQA rankings diverge from traditional VQA benchmarks while correlating strongly with SpatialWorld, suggesting that conventional benchmarks do not fully capture agentic readiness and that AgentVQA offers a reliable, substantially lower-cost alternative for rapid agent-model evaluation.

Taken together, these findings support viewing agentic competence as a collection of domain-specific capabilities rather than a single general ability. The strong agreement with online evaluation in the matched domain, alongside weaker cross-domain agreement, suggests that AgentVQA can help distinguish these capabilities and guide more targeted model development. Within the broader evaluation landscape, controlled offline benchmarks can complement interactive environments by identifying weaknesses efficiently and directing subsequent online testing toward the domains and failure modes that require closer investigation.

Future work should connect specific offline error patterns to failures during interactive execution. Such analyses could guide improvements in grounding and state tracking while testing whether gains on controlled questions translate into more reliable online behavior.

\section{Acknowledgements}

This work was supported by Princeton Research Computing resources, including the Neuronic high-performance computing cluster. We thank Princeton Language and Intelligence (PLI) for their support of this research.

{
    \small
    \bibliographystyle{ieeenat_fullname}
    \bibliography{main}

\begin{thebibliography}{82}
\providecommand{\natexlab}[1]{#1}
\providecommand{\url}[1]{\texttt{#1}}
\expandafter\ifx\csname urlstyle\endcsname\relax
  \providecommand{\doi}[1]{doi: #1}\else
  \providecommand{\doi}{doi: \begingroup \urlstyle{rm}\Url}\fi

\bibitem[Anthropic(2026{\natexlab{a}})]{opus}
Anthropic.
\newblock {Claude Opus 4.8 System Card}.
\newblock \url{https://www-cdn.anthropic.com/0b4915911bb0d19eca5b5ee635c80fef830a37ea.pdf}, 2026{\natexlab{a}}.

\bibitem[Anthropic(2026{\natexlab{b}})]{sonnet}
Anthropic.
\newblock {Claude Sonnet 5 System Card}.
\newblock \url{https://www-cdn.anthropic.com/283ef97c476cf442c91d9a37d5b214242a55bb92/Claude%20Sonnet%205%20System%20Card.pdf}, 2026{\natexlab{b}}.

\bibitem[Bai et~al.(2023)Bai, Bai, Chu, Cui, Dang, Deng, Fan, Ge, Han, Huang, et~al.]{bai2023qwenvlversatilevisionlanguagemodel}
Jinze Bai, Shuai Bai, Yunfei Chu, Zeyu Cui, Kai Dang, Xiaodong Deng, Yang Fan, Wenbin Ge, Yu Han, Fei Huang, et~al.
\newblock Qwen technical report.
\newblock \emph{arXiv preprint arXiv:2309.16609}, 2023.

\bibitem[Balunovi{\'c} et~al.(2025)Balunovi{\'c}, Dekoninck, Petrov, Jovanovi{\'c}, and Vechev]{balunovic_srimatharena_2025}
Mislav Balunovi{\'c}, Jasper Dekoninck, Ivo Petrov, Nikola Jovanovi{\'c}, and Martin Vechev.
\newblock Matharena: Evaluating llms on uncontaminated math competitions.
\newblock \emph{arXiv preprint arXiv:2505.23281}, 2025.

\bibitem[Bengio et~al.(2025)Bengio, Cohen, Fornasiere, Ghosn, Greiner, MacDermott, Mindermann, Oberman, Richardson, Richardson, Rondeau, St-Charles, and Williams-King]{bengio2025superintelligentagentsposecatastrophic}
Yoshua Bengio, Michael Cohen, Damiano Fornasiere, Joumana Ghosn, Pietro Greiner, Matt MacDermott, Sören Mindermann, Adam Oberman, Jesse Richardson, Oliver Richardson, Marc-Antoine Rondeau, Pierre-Luc St-Charles, and David Williams-King.
\newblock Superintelligent agents pose catastrophic risks: Can scientist ai offer a safer path?, 2025.

\bibitem[Black et~al.(2024)Black, Brown, Driess, Esmail, Equi, Finn, Fusai, Groom, Hausman, Ichter, et~al.]{black2410pi0}
Kevin Black, Noah Brown, Danny Driess, Adnan Esmail, Michael Equi, Chelsea Finn, Niccolo Fusai, Lachy Groom, Karol Hausman, Brian Ichter, et~al.
\newblock $\pi_0 $: A vision-language-action flow model for general robot control.
\newblock \emph{arXiv preprint arXiv:2410.24164}, 2024.

\bibitem[Black et~al.(2025)Black, Brown, Darpinian, Dhabalia, Driess, Esmail, Equi, Finn, Fusai, Galliker, Ghosh, Groom, Hausman, ichter, Jakubczak, Jones, Ke, LeBlanc, Levine, Li-Bell, Mothukuri, Nair, Pertsch, Ren, Shi, Smith, Springenberg, Stachowicz, Tanner, Vuong, Walke, Walling, Wang, Yu, and Zhilinsky]{pi0.5}
Kevin Black, Noah Brown, James Darpinian, Karan Dhabalia, Danny Driess, Adnan Esmail, Michael~Robert Equi, Chelsea Finn, Niccolo Fusai, Manuel~Y. Galliker, Dibya Ghosh, Lachy Groom, Karol Hausman, brian ichter, Szymon Jakubczak, Tim Jones, Liyiming Ke, Devin LeBlanc, Sergey Levine, Adrian Li-Bell, Mohith Mothukuri, Suraj Nair, Karl Pertsch, Allen~Z. Ren, Lucy~Xiaoyang Shi, Laura Smith, Jost~Tobias Springenberg, Kyle Stachowicz, James Tanner, Quan Vuong, Homer Walke, Anna Walling, Haohuan Wang, Lili Yu, and Ury Zhilinsky.
\newblock $\pi_{0.5}$: a vision-language-action model with open-world generalization.
\newblock In \emph{Proceedings of The 9th Conference on Robot Learning}, pages 17--40. PMLR, 2025.

\bibitem[Black et~al.(2026)Black, Brown, Driess, Esmail, Equi, Finn, Fusai, Groom, Hausman, Ichter, Jakubczak, Jones, Ke, Levine, Li-Bell, Mothukuri, Nair, Pertsch, Shi, Tanner, Vuong, Walling, Wang, and Zhilinsky]{pi0}
Kevin Black, Noah Brown, Danny Driess, Adnan Esmail, Michael Equi, Chelsea Finn, Niccolo Fusai, Lachy Groom, Karol Hausman, Brian Ichter, Szymon Jakubczak, Tim Jones, Liyiming Ke, Sergey Levine, Adrian Li-Bell, Mohith Mothukuri, Suraj Nair, Karl Pertsch, Lucy~Xiaoyang Shi, James Tanner, Quan Vuong, Anna Walling, Haohuan Wang, and Ury Zhilinsky.
\newblock $\pi_0$: A vision-language-action flow model for general robot control, 2026.

\bibitem[Bonatti et~al.(2025)Bonatti, Zhao, Bonacci, Dupont, Abdali, Li, Lu, Wagle, Koishida, Bucker, Jang, and Hui]{bonatti2025windows}
Rogerio Bonatti, Dan Zhao, Francesco Bonacci, Dillon Dupont, Sara Abdali, Yinheng Li, Yadong Lu, Justin Wagle, Kazuhito Koishida, Arthur Bucker, Lawrence~Keunho Jang, and Zheng Hui.
\newblock Windows agent arena: Evaluating multi-modal {OS} agents at scale.
\newblock In \emph{Forty-second International Conference on Machine Learning}, 2025.

\bibitem[Chen et~al.(2025)Chen, Xie, Ma, Sanketi, and Goldberg]{chen2025robo2vlm}
Kaiyuan Chen, Shuangyu Xie, Zehan Ma, Pannag~R Sanketi, and Ken Goldberg.
\newblock Robo2vlm: Visual question answering from large-scale in-the-wild robot manipulation datasets.
\newblock \emph{arXiv preprint arXiv:2505.15517}, 2025.

\bibitem[Chen et~al.(2023)Chen, Ge, Ge, Ding, Li, Wang, Xu, Shan, and Liu]{chen2024egoplanbenchbenchmarkingmultimodallarge}
Yi Chen, Yuying Ge, Yixiao Ge, Mingyu Ding, Bohao Li, Rui Wang, Ruifeng Xu, Ying Shan, and Xihui Liu.
\newblock Egoplan-bench: Benchmarking multimodal large language models for human-level planning.
\newblock \emph{arXiv preprint arXiv:2312.06722}, 2023.

\bibitem[Cheng et~al.(2024{\natexlab{a}})Cheng, Sun, Chu, Xu, YanTao, Zhang, and Wu]{cheng2024seeclick}
Kanzhi Cheng, Qiushi Sun, Yougang Chu, Fangzhi Xu, Li YanTao, Jianbing Zhang, and Zhiyong Wu.
\newblock {S}ee{C}lick: Harnessing {GUI} grounding for advanced visual {GUI} agents.
\newblock In \emph{ACL}, 2024{\natexlab{a}}.

\bibitem[Cheng et~al.(2024{\natexlab{b}})Cheng, Fang, Yu, Zhou, Li, Tian, Li, Han, and Liu]{cheng2024videgothinkassessingegocentricvideo}
Sijie Cheng, Kechen Fang, Yangyang Yu, Sicheng Zhou, Bohao Li, Ye Tian, Tingguang Li, Lei Han, and Yang Liu.
\newblock Videgothink: Assessing egocentric video understanding capabilities for embodied ai.
\newblock \emph{arXiv preprint arXiv:2410.11623}, 2024{\natexlab{b}}.

\bibitem[Cheng et~al.(2024{\natexlab{c}})Cheng, Guo, Wu, Fang, Li, Liu, and Liu]{cheng2024egothinkevaluatingfirstpersonperspective}
Sijie Cheng, Zhicheng Guo, Jingwen Wu, Kechen Fang, Peng Li, Huaping Liu, and Yang Liu.
\newblock Egothink: Evaluating first-person perspective thinking capability of vision-language models.
\newblock In \emph{CVPR}, 2024{\natexlab{c}}.

\bibitem[Dasari et~al.(2022)Dasari, Wang, Hong, Bahl, Lin, Wang, Thankaraj, Chahal, Calli, Gupta, et~al.]{dasari2022rb2}
Sudeep Dasari, Jianren Wang, Joyce Hong, Shikhar Bahl, Yixin Lin, Austin Wang, Abitha Thankaraj, Karanbir Chahal, Berk Calli, Saurabh Gupta, et~al.
\newblock Rb2: Robotic manipulation benchmarking with a twist.
\newblock \emph{arXiv preprint arXiv:2203.08098}, 2022.

\bibitem[Deepmind(2026)]{comanici2025gemini_base}
Google Deepmind.
\newblock {Gemini 3.7 Flash System Card}.
\newblock \url{https://deepmind.google/models/model-cards/gemini-3-7-flash/}, 2026.

\bibitem[Du et~al.(2024)Du, Wu, Li, Huang, and Wei]{du2024embspatialbenchbenchmarkingspatialunderstanding}
Mengfei Du, Binhao Wu, Zejun Li, Xuanjing Huang, and Zhongyu Wei.
\newblock Embspatial-bench: Benchmarking spatial understanding for embodied tasks with large vision-language models.
\newblock \emph{arXiv preprint arXiv:2406.05756}, 2024.

\bibitem[Fan et~al.(2022)Fan, Wang, Jiang, Mandlekar, Yang, Zhu, Tang, Huang, Zhu, and Anandkumar]{fan2022minedojobuildingopenendedembodied}
Linxi Fan, Guanzhi Wang, Yunfan Jiang, Ajay Mandlekar, Yuncong Yang, Haoyi Zhu, Andrew Tang, De-An Huang, Yuke Zhu, and Anima Anandkumar.
\newblock Minedojo: Building open-ended embodied agents with internet-scale knowledge.
\newblock \emph{Advances in Neural Information Processing Systems}, 35, 2022.

\bibitem[Gao et~al.(2026)Gao, Qu, Tang, Wang, Huang, Qiao, Huang, Yang, Li, Yuan, Li, Zeng, Li, Wang, Liu, Huang, Huang, Zhang, Huang, Duan, and Dong]{gao2026spatialworldbenchmarkinginteractivespatial}
Hongcheng Gao, Hailong Qu, Jingyi Tang, Jiahao Wang, Zihao Huang, Hengkang Qiao, Shihong Huang, Junming Yang, Yi Li, Hongyixuan Yuan, Wenjie Li, Bohan Zeng, Wenbo Li, Bo Wang, Jianhui Liu, Olive Huang, Haoyang Huang, Wentao Zhang, Guoqing Huang, Nan Duan, and Yinpeng Dong.
\newblock Spatialworld: Benchmarking interactive spatial reasoning of multimodal agents in real-world tasks, 2026.

\bibitem[{GLM-V Team}(2026)]{vteam2026glm45vglm41vthinkingversatilemultimodal}
{GLM-V Team}.
\newblock Glm-4.5v and glm-4.1v-thinking: Towards versatile multimodal reasoning with scalable reinforcement learning, 2026.

\bibitem[Gong et~al.(2025)Gong, Li, Ma, Li, Ji, Yang, Luo, Yan, and Ji]{gong2025space10comprehensivebenchmarkmultimodal}
Ziyang Gong, Wenhao Li, Oliver Ma, Songyuan Li, Jiayi Ji, Xue Yang, Gen Luo, Junchi Yan, and Rongrong Ji.
\newblock Space-10: A comprehensive benchmark for multimodal large language models in compositional spatial intelligence.
\newblock \emph{arXiv preprint arXiv:2506.07966}, 2025.

\bibitem[Google(2026)]{gemini3}
Google.
\newblock {Gemini 3.1 Pro Model Card}.
\newblock \url{https://storage.googleapis.com/deepmind-media/Model-Cards/Gemini-3-1-Pro-Model-Card.pdf}, 2026.

\bibitem[Grauman et~al.(2024)Grauman, Westbury, Torresani, Kitani, Malik, Afouras, Ashutosh, Baiyya, Bansal, Boote, et~al.]{grauman2024egoexo4dunderstandingskilledhuman}
Kristen Grauman, Andrew Westbury, Lorenzo Torresani, Kris Kitani, Jitendra Malik, Triantafyllos Afouras, Kumar Ashutosh, Vijay Baiyya, Siddhant Bansal, Bikram Boote, et~al.
\newblock Ego-exo4d: Understanding skilled human activity from first-and third-person perspectives.
\newblock In \emph{CVPR}, 2024.

\bibitem[Hafner(2021)]{hafner2022benchmarkingspectrumagentcapabilities}
Danijar Hafner.
\newblock Benchmarking the spectrum of agent capabilities.
\newblock \emph{arXiv preprint arXiv:2109.06780}, 2021.

\bibitem[Henderson et~al.(2018)Henderson, Islam, Bachman, Pineau, Precup, and Meger]{henderson2018deep}
Peter Henderson, Riashat Islam, Philip Bachman, Joelle Pineau, Doina Precup, and David Meger.
\newblock Deep reinforcement learning that matters.
\newblock In \emph{AAAI}, 2018.

\bibitem[Hurst et~al.(2024)Hurst, Lerer, Goucher, Perelman, Ramesh, Clark, Ostrow, Welihinda, Hayes, Radford, et~al.]{openai2024gpt4ocard}
Aaron Hurst, Adam Lerer, Adam~P Goucher, Adam Perelman, Aditya Ramesh, Aidan Clark, AJ Ostrow, Akila Welihinda, Alan Hayes, Alec Radford, et~al.
\newblock Gpt-4o system card.
\newblock \emph{arXiv preprint arXiv:2410.21276}, 2024.

\bibitem[Jang et~al.(2024)Jang, Li, Zhao, Ding, Lin, Liang, Bonatti, and Koishida]{jang2025videowebarenaevaluatinglongcontext}
Lawrence Jang, Yinheng Li, Dan Zhao, Charles Ding, Justin Lin, Paul~Pu Liang, Rogerio Bonatti, and Kazuhito Koishida.
\newblock Videowebarena: Evaluating long context multimodal agents with video understanding web tasks.
\newblock \emph{arXiv preprint arXiv:2410.19100}, 2024.

\bibitem[Jang et~al.(2025)Jang, Song, Sohn, Logeswaran, Luo, Kim, Bae, and Lee]{jang2025scalablevideotodatasetgenerationcrossplatform}
Yunseok Jang, Yeda Song, Sungryull Sohn, Lajanugen Logeswaran, Tiange Luo, Dong-Ki Kim, Kyunghoon Bae, and Honglak Lee.
\newblock Scalable video-to-dataset generation for cross-platform mobile agents.
\newblock In \emph{CVPR}, 2025.

\bibitem[Ji et~al.(2025)Ji, Tan, Shi, Hao, Zhang, Zhang, Wang, Zhao, Mu, An, Xue, Su, Lyu, Zheng, Liu, Wang, and Zhang]{ji2025robobrainunifiedbrainmodel}
Yuheng Ji, Huajie Tan, Jiayu Shi, Xiaoshuai Hao, Yuan Zhang, Hengyuan Zhang, Pengwei Wang, Mengdi Zhao, Yao Mu, Pengju An, Xinda Xue, Qinghang Su, Huaihai Lyu, Xiaolong Zheng, Jiaming Liu, Zhongyuan Wang, and Shanghang Zhang.
\newblock Robobrain: A unified brain model for robotic manipulation from abstract to concrete, 2025.

\bibitem[Jia et~al.(2026)Jia, Qi, Zhang, Zhang, Yu, He, Wang, and Yi]{jia2026omnispatialcomprehensivespatialreasoning}
Mengdi Jia, Zekun Qi, Shaochen Zhang, Wenyao Zhang, Xinqiang Yu, Jiawei He, He Wang, and Li Yi.
\newblock Omnispatial: Towards comprehensive spatial reasoning benchmark for vision language models, 2026.

\bibitem[Jiang et~al.(2022)Jiang, Gupta, Zhang, Wang, Dou, Chen, Fei-Fei, Anandkumar, Zhu, and Fan]{jiang2023vimageneralrobotmanipulation}
Yunfan Jiang, Agrim Gupta, Zichen Zhang, Guanzhi Wang, Yongqiang Dou, Yanjun Chen, Li Fei-Fei, Anima Anandkumar, Yuke Zhu, and Linxi Fan.
\newblock Vima: General robot manipulation with multimodal prompts.
\newblock \emph{arXiv preprint arXiv:2210.03094}, 2\penalty0 (3), 2022.

\bibitem[Jimenez et~al.(2023)Jimenez, Yang, Wettig, Yao, Pei, Press, and Narasimhan]{jimenez2023swe}
Carlos~E Jimenez, John Yang, Alexander Wettig, Shunyu Yao, Kexin Pei, Ofir Press, and Karthik Narasimhan.
\newblock Swe-bench: Can language models resolve real-world github issues?
\newblock \emph{arXiv preprint arXiv:2310.06770}, 2023.

\bibitem[Koh et~al.(2024)Koh, Lo, Jang, Duvvur, Lim, Huang, Neubig, Zhou, Salakhutdinov, and Fried]{koh2024visualwebarena}
Jing~Yu Koh, Robert Lo, Lawrence Jang, Vikram Duvvur, Ming~Chong Lim, Po-Yu Huang, Graham Neubig, Shuyan Zhou, Ruslan Salakhutdinov, and Daniel Fried.
\newblock Visualwebarena: Evaluating multimodal agents on realistic visual web tasks.
\newblock \emph{arXiv preprint arXiv:2401.13649}, 2024.

\bibitem[Li et~al.(2023{\natexlab{a}})Li, Wang, Wang, Ge, Ge, and Shan]{li2023seed}
Bohao Li, Rui Wang, Guangzhi Wang, Yuying Ge, Yixiao Ge, and Ying Shan.
\newblock Seed-bench: Benchmarking multimodal llms with generative comprehension.
\newblock \emph{arXiv preprint arXiv:2307.16125}, 2023{\natexlab{a}}.

\bibitem[Li et~al.(2023{\natexlab{b}})Li, Zhang, Wong, Gokmen, Srivastava, Mart{\'\i}n-Mart{\'\i}n, Wang, Levine, Lingelbach, Sun, et~al.]{li2024behavior1khumancenteredembodiedai}
Chengshu Li, Ruohan Zhang, Josiah Wong, Cem Gokmen, Sanjana Srivastava, Roberto Mart{\'\i}n-Mart{\'\i}n, Chen Wang, Gabrael Levine, Michael Lingelbach, Jiankai Sun, et~al.
\newblock Behavior-1k: A benchmark for embodied ai with 1,000 everyday activities and realistic simulation.
\newblock In \emph{CoRL}. PMLR, 2023{\natexlab{b}}.

\bibitem[Li et~al.(2025)Li, Meng, Lin, Luo, Tian, Ma, Huang, and Chua]{li2025screenspotproguigroundingprofessional}
Kaixin Li, Ziyang Meng, Hongzhan Lin, Ziyang Luo, Yuchen Tian, Jing Ma, Zhiyong Huang, and Tat-Seng Chua.
\newblock Screenspot-pro: Gui grounding for professional high-resolution computer use.
\newblock \emph{arXiv preprint arXiv:2504.07981}, 2025.

\bibitem[Liu et~al.(2023)Liu, Yu, Zhang, Xu, Lei, Lai, Gu, Ding, Men, Yang, et~al.]{liu2023agentbenchevaluatingllmsagents}
Xiao Liu, Hao Yu, Hanchen Zhang, Yifan Xu, Xuanyu Lei, Hanyu Lai, Yu Gu, Hangliang Ding, Kaiwen Men, Kejuan Yang, et~al.
\newblock Agentbench: Evaluating llms as agents.
\newblock \emph{arXiv preprint arXiv:2308.03688}, 2023.

\bibitem[Liu et~al.(2024)Liu, Duan, Zhang, Li, Zhang, Zhao, Yuan, Wang, He, Liu, et~al.]{liu2024mmbench}
Yuan Liu, Haodong Duan, Yuanhan Zhang, Bo Li, Songyang Zhang, Wangbo Zhao, Yike Yuan, Jiaqi Wang, Conghui He, Ziwei Liu, et~al.
\newblock Mmbench: Is your multi-modal model an all-around player?
\newblock In \emph{ECCV}. Springer, 2024.

\bibitem[Lu et~al.(2023)Lu, Fan, Deng, Liu, Li, and Wang]{lu2023vlgrasp6dofinteractivegrasp}
Yuhao Lu, Yixuan Fan, Beixing Deng, Fangfu Liu, Yali Li, and Shengjin Wang.
\newblock Vl-grasp: a 6-dof interactive grasp policy for language-oriented objects in cluttered indoor scenes.
\newblock In \emph{IROS}. IEEE, 2023.

\bibitem[Ma et~al.(2024)Ma, Chen, Zhang, Chou, de~Melo, and Yuille]{ma20253dsrbenchcomprehensive3dspatial}
Wufei Ma, Haoyu Chen, Guofeng Zhang, Yu-Cheng Chou, Celso~M de Melo, and Alan Yuille.
\newblock 3dsrbench: A comprehensive 3d spatial reasoning benchmark.
\newblock \emph{arXiv preprint arXiv:2412.07825}, 2024.

\bibitem[Majumdar et~al.(2024)Majumdar, Ajay, Zhang, Putta, Yenamandra, Henaff, Silwal, Mcvay, Maksymets, Arnaud, et~al.]{majumdar2024openeqa}
Arjun Majumdar, Anurag Ajay, Xiaohan Zhang, Pranav Putta, Sriram Yenamandra, Mikael Henaff, Sneha Silwal, Paul Mcvay, Oleksandr Maksymets, Sergio Arnaud, et~al.
\newblock Openeqa: Embodied question answering in the era of foundation models.
\newblock In \emph{CVPR}, 2024.

\bibitem[Masry et~al.(2025)Masry, Islam, Ahmed, Bajaj, Kabir, Kartha, Laskar, Rahman, Rahman, Shahmohammadi, Thakkar, Parvez, Hoque, and Joty]{masry2025chartqaprodiversechallengingbenchmark}
Ahmed Masry, Mohammed~Saidul Islam, Mahir Ahmed, Aayush Bajaj, Firoz Kabir, Aaryaman Kartha, Md~Tahmid~Rahman Laskar, Mizanur Rahman, Shadikur Rahman, Mehrad Shahmohammadi, Megh Thakkar, Md~Rizwan Parvez, Enamul Hoque, and Shafiq Joty.
\newblock Chartqapro: A more diverse and challenging benchmark for chart question answering, 2025.

\bibitem[Mees et~al.(2022)Mees, Hermann, Rosete-Beas, and Burgard]{mees2022calvinbenchmarklanguageconditionedpolicy}
Oier Mees, Lukas Hermann, Erick Rosete-Beas, and Wolfram Burgard.
\newblock Calvin: A benchmark for language-conditioned policy learning for long-horizon robot manipulation tasks.
\newblock \emph{IEEE Robotics and Automation Letters}, 7\penalty0 (3), 2022.

\bibitem[Meta(2025)]{meta2024llama4}
Meta.
\newblock {The Llama 4 herd: The beginning of a new era of natively multimodal AI innovation}.
\newblock \url{https://ai.meta.com/blog/llama-4-multimodal-intelligence/}, 2025.

\bibitem[Milani et~al.(2023)Milani, Kanervisto, Ramanauskas, Schulhoff, Houghton, and Shah]{milani2023beddminerlbasaltevaluation}
Stephanie Milani, Anssi Kanervisto, Karolis Ramanauskas, Sander Schulhoff, Brandon Houghton, and Rohin Shah.
\newblock Bedd: The minerl basalt evaluation and demonstrations dataset for training and benchmarking agents that solve fuzzy tasks.
\newblock \emph{Advances in Neural Information Processing Systems}, 36:\penalty0 32867--32878, 2023.

\bibitem[Mun et~al.(2017)Mun, Hongsuck~Seo, Jung, and Han]{mun2017marioqaansweringquestionswatching}
Jonghwan Mun, Paul Hongsuck~Seo, Ilchae Jung, and Bohyung Han.
\newblock Marioqa: Answering questions by watching gameplay videos.
\newblock In \emph{ICCV}, 2017.

\bibitem[Nasiriany et~al.(2024)Nasiriany, Maddukuri, Zhang, Parikh, Lo, Joshi, Mandlekar, and Zhu]{nasiriany2024robocasalargescalesimulationeveryday}
Soroush Nasiriany, Abhiram Maddukuri, Lance Zhang, Adeet Parikh, Aaron Lo, Abhishek Joshi, Ajay Mandlekar, and Yuke Zhu.
\newblock Robocasa: Large-scale simulation of everyday tasks for generalist robots.
\newblock \emph{arXiv preprint arXiv:2406.02523}, 2024.

\bibitem[OpenAI(2026)]{openai2025gpt5}
OpenAI.
\newblock {GPT-5.6 System Card}.
\newblock \url{https://deploymentsafety.openai.com/gpt-5-6}, 2026.

\bibitem[Paglieri et~al.(2024)Paglieri, Cupia{\l}, Coward, Piterbarg, Wolczyk, Khan, Pignatelli, Kuci{\'n}ski, Pinto, Fergus, et~al.]{paglieri2025balrogbenchmarkingagenticllm}
Davide Paglieri, Bart{\l}omiej Cupia{\l}, Samuel Coward, Ulyana Piterbarg, Maciej Wolczyk, Akbir Khan, Eduardo Pignatelli, {\L}ukasz Kuci{\'n}ski, Lerrel Pinto, Rob Fergus, et~al.
\newblock Balrog: Benchmarking agentic llm and vlm reasoning on games.
\newblock \emph{arXiv preprint arXiv:2411.13543}, 2024.

\bibitem[Patraucean et~al.(2023)Patraucean, Smaira, Gupta, Recasens, Markeeva, Banarse, Koppula, Malinowski, Yang, Doersch, et~al.]{patraucean2023perceptiontestdiagnosticbenchmark}
Viorica Patraucean, Lucas Smaira, Ankush Gupta, Adria Recasens, Larisa Markeeva, Dylan Banarse, Skanda Koppula, Mateusz Malinowski, Yi Yang, Carl Doersch, et~al.
\newblock Perception test: A diagnostic benchmark for multimodal video models.
\newblock \emph{Advances in Neural Information Processing Systems}, 36, 2023.

\bibitem[Plaat et~al.(2025)Plaat, van Duijn, van Stein, Preuss, van~der Putten, and Batenburg]{plaat2025agenticlargelanguagemodels}
Aske Plaat, Max van Duijn, Niki van Stein, Mike Preuss, Peter van~der Putten, and Kees~Joost Batenburg.
\newblock Agentic large language models, a survey, 2025.

\bibitem[{Qwen Team}(2026{\natexlab{a}})]{qwen35}
{Qwen Team}.
\newblock Qwen3.5.
\newblock \url{https://qwen.ai/blog?id=qwen3.5}, 2026{\natexlab{a}}.

\bibitem[{Qwen Team}(2026{\natexlab{b}})]{qwen37}
{Qwen Team}.
\newblock Qwen3.7.
\newblock \url{https://qwen.ai/blog?id=qwen3.7}, 2026{\natexlab{b}}.

\bibitem[{Qwen Team}(2026{\natexlab{c}})]{qwen38}
{Qwen Team}.
\newblock Qwen3.8.
\newblock \url{https://qwen.ai/blog?id=qwen3.8}, 2026{\natexlab{c}}.

\bibitem[Rawles et~al.(2023)Rawles, Li, Rodriguez, Riva, and Lillicrap]{rawles2023androidwildlargescaledataset}
Christopher Rawles, Alice Li, Daniel Rodriguez, Oriana Riva, and Timothy Lillicrap.
\newblock Androidinthewild: A large-scale dataset for android device control.
\newblock \emph{Advances in Neural Information Processing Systems}, 36, 2023.

\bibitem[Rawles et~al.(2024)Rawles, Clinckemaillie, Chang, Waltz, Lau, Fair, Li, Bishop, Li, Campbell-Ajala, et~al.]{rawles2024androidworld}
Christopher Rawles, Sarah Clinckemaillie, Yifan Chang, Jonathan Waltz, Gabrielle Lau, Marybeth Fair, Alice Li, William Bishop, Wei Li, Folawiyo Campbell-Ajala, et~al.
\newblock Androidworld: A dynamic benchmarking environment for autonomous agents.
\newblock \emph{arXiv preprint arXiv:2405.14573}, 2024.

\bibitem[Reed et~al.(2022)Reed, Zolna, Parisotto, Colmenarejo, Novikov, Barth-Maron, Gimenez, Sulsky, Kay, Springenberg, Eccles, Bruce, Razavi, Edwards, Heess, Chen, Hadsell, Vinyals, Bordbar, and de~Freitas]{reed2022generalistagent}
Scott Reed, Konrad Zolna, Emilio Parisotto, Sergio~Gomez Colmenarejo, Alexander Novikov, Gabriel Barth-Maron, Mai Gimenez, Yury Sulsky, Jackie Kay, Jost~Tobias Springenberg, Tom Eccles, Jake Bruce, Ali Razavi, Ashley Edwards, Nicolas Heess, Yutian Chen, Raia Hadsell, Oriol Vinyals, Mahyar Bordbar, and Nando de Freitas.
\newblock A generalist agent, 2022.

\bibitem[Rein et~al.(2024)Rein, Hou, Stickland, Petty, Pang, Dirani, Michael, and Bowman]{rein2024gpqa}
David Rein, Betty~Li Hou, Asa~Cooper Stickland, Jackson Petty, Richard~Yuanzhe Pang, Julien Dirani, Julian Michael, and Samuel~R Bowman.
\newblock Gpqa: A graduate-level google-proof q\&a benchmark.
\newblock In \emph{COLM}, 2024.

\bibitem[Sarch et~al.(2025)Sarch, Kumaravel, Ravi, Vineet, and Wilson]{sarch2025grounding}
Gabriel Sarch, Balasaravanan~Thoravi Kumaravel, Sahithya Ravi, Vibhav Vineet, and Andrew~D Wilson.
\newblock Grounding task assistance with multimodal cues from a single demonstration.
\newblock \emph{arXiv preprint arXiv:2505.01578}, 2025.

\bibitem[Savva et~al.(2019)Savva, Kadian, Maksymets, Zhao, Wijmans, Jain, Straub, Liu, Koltun, Malik, et~al.]{savva2019habitatplatformembodiedai}
Manolis Savva, Abhishek Kadian, Oleksandr Maksymets, Yili Zhao, Erik Wijmans, Bhavana Jain, Julian Straub, Jia Liu, Vladlen Koltun, Jitendra Malik, et~al.
\newblock Habitat: A platform for embodied ai research.
\newblock In \emph{ICCV}, 2019.

\bibitem[Shiri et~al.(2024)Shiri, Guo, Far, Yu, Haffari, and Li]{shiri2024empiricalanalysisspatialreasoning}
Fatemeh Shiri, Xiao-Yu Guo, Mona~Golestan Far, Xin Yu, Gholamreza Haffari, and Yuan-Fang Li.
\newblock An empirical analysis on spatial reasoning capabilities of large multimodal models, 2024.

\bibitem[Sutton et~al.(1998)Sutton, Barto, et~al.]{sutton1998reinforcement}
Richard~S Sutton, Andrew~G Barto, et~al.
\newblock \emph{Reinforcement learning: An introduction}.
\newblock MIT press Cambridge, 1998.

\bibitem[Taesiri and Bezemer(2024)]{taesiri2024videogamebunnyvisionassistantsvideo}
Mohammad~Reza Taesiri and Cor-Paul Bezemer.
\newblock Videogamebunny: Towards vision assistants for video games, 2024.

\bibitem[Team et~al.(2025)Team, Abeyruwan, Ainslie, Alayrac, Arenas, Armstrong, Balakrishna, Baruch, Bauza, Blokzijl, et~al.]{comanici2025geminirobotics}
Gemini~Robotics Team, Saminda Abeyruwan, Joshua Ainslie, Jean-Baptiste Alayrac, Montserrat~Gonzalez Arenas, Travis Armstrong, Ashwin Balakrishna, Robert Baruch, Maria Bauza, Michiel Blokzijl, et~al.
\newblock Gemini robotics: Bringing ai into the physical world.
\newblock \emph{arXiv preprint arXiv:2503.20020}, 2025.

\bibitem[Team(2026)]{kimi}
Kimi Team.
\newblock Kimi k3: Open frontier intelligence, 2026.

\bibitem[Tong et~al.(2025)Tong, Tang, Li, Mou, Zhang, Zhao, Wen, Song, Zhan, Lu, Tao, Guo, Yu, Cheng, Jiang, Wang, Liang, Fei, Wan, Ma, Ge, Chen, Gui, Qiu, Zhang, and Huang]{tong2025gamerlsynthesizingverifiablegame}
Jingqi Tong, Jixin Tang, Hangcheng Li, Yurong Mou, Ming Zhang, Jun Zhao, Yanbo Wen, Fan Song, Jiahao Zhan, Yuyang Lu, Chaoran Tao, Zhiyuan Guo, Jizhou Yu, Tianhao Cheng, Changhao Jiang, Zhen Wang, Tao Liang, Zhihui Fei, Mingyang Wan, Guojun Ma, Weifeng Ge, Guanhua Chen, Tao Gui, Xipeng Qiu, Qi Zhang, and Xuanjing Huang.
\newblock Game-rl: Synthesizing verifiable game tasks at scale to boost vlms general reasoning.
\newblock \emph{arXiv preprint arXiv:2505.13886}, 2025.

\bibitem[Vuong et~al.(2023)Vuong, Levine, Walke, Pertsch, Singh, Doshi, Xu, Luo, Tan, Shah, et~al.]{embodimentcollaboration2025openxembodimentroboticlearning}
Quan Vuong, Sergey Levine, Homer~Rich Walke, Karl Pertsch, Anikait Singh, Ria Doshi, Charles Xu, Jianlan Luo, Liam Tan, Dhruv Shah, et~al.
\newblock Open x-embodiment: Robotic learning datasets and rt-x models.
\newblock In \emph{CoRL}, 2023.

\bibitem[Wang et~al.(2023)Wang, Kwon, Rad, Pan, Chakraborty, Andrist, Bohus, Feniello, Tekin, Frujeri, et~al.]{wang2023holoassistegocentrichumaninteraction}
Xin Wang, Taein Kwon, Mahdi Rad, Bowen Pan, Ishani Chakraborty, Sean Andrist, Dan Bohus, Ashley Feniello, Bugra Tekin, Felipe~Vieira Frujeri, et~al.
\newblock Holoassist: an egocentric human interaction dataset for interactive ai assistants in the real world.
\newblock In \emph{ICCV}, 2023.

\bibitem[Wong et~al.(2025)Wong, Wang, Zhao, Chen, Gao, Zhang, Zhou, Wang, Xiang, Zhang, et~al.]{wong2025widesearchbenchmarkingagenticbroad}
Ryan Wong, Jiawei Wang, Junjie Zhao, Li Chen, Yan Gao, Long Zhang, Xuan Zhou, Zuo Wang, Kai Xiang, Ge Zhang, et~al.
\newblock Widesearch: Benchmarking agentic broad info-seeking.
\newblock \emph{arXiv preprint arXiv:2508.07999}, 2025.

\bibitem[Xu et~al.(2024)Xu, Song, Li, Tang, Jain, Bao, Wang, Zhou, Guo, Cao, et~al.]{xu2025theagentcompanybenchmarkingllmagents}
Frank~F Xu, Yufan Song, Boxuan Li, Yuxuan Tang, Kritanjali Jain, Mengxue Bao, Zora~Z Wang, Xuhui Zhou, Zhitong Guo, Murong Cao, et~al.
\newblock Theagentcompany: benchmarking llm agents on consequential real world tasks.
\newblock \emph{arXiv preprint arXiv:2412.14161}, 2024.

\bibitem[Yang et~al.(2025{\natexlab{a}})Yang, Yang, Gupta, Han, Fei-Fei, and Xie]{yang2025thinking}
Jihan Yang, Shusheng Yang, Anjali~W Gupta, Rilyn Han, Li Fei-Fei, and Saining Xie.
\newblock Thinking in space: How multimodal large language models see, remember, and recall spaces.
\newblock In \emph{CVPR}, 2025{\natexlab{a}}.

\bibitem[Yang et~al.(2025{\natexlab{b}})Yang, Chen, Zhang, Zhao, Qian, Wang, Wang, Koripella, Movahedi, Li, Ji, Zhang, and Zhang]{pmlr-v267-yang25f}
Rui Yang, Hanyang Chen, Junyu Zhang, Mark Zhao, Cheng Qian, Kangrui Wang, Qineng Wang, Teja~Venkat Koripella, Marziyeh Movahedi, Manling Li, Heng Ji, Huan Zhang, and Tong Zhang.
\newblock {E}mbodied{B}ench: Comprehensive benchmarking multi-modal large language models for vision-driven embodied agents.
\newblock In \emph{Proceedings of the 42nd International Conference on Machine Learning}. PMLR, 2025{\natexlab{b}}.

\bibitem[Yao et~al.(2024)Yao, Shinn, Razavi, and Narasimhan]{yao2024taubenchbenchmarktoolagentuserinteraction}
Shunyu Yao, Noah Shinn, Pedram Razavi, and Karthik Narasimhan.
\newblock Tau-bench: A benchmark for tool-agent-user interaction in real-world domains.
\newblock \emph{arXiv preprint arXiv:2406.12045}, 2024.

\bibitem[Yehudai et~al.(2025)Yehudai, Eden, Li, Uziel, Zhao, Bar-Haim, Cohan, and Shmueli-Scheuer]{yehudai2025surveyevaluationllmbasedagents}
Asaf Yehudai, Lilach Eden, Alan Li, Guy Uziel, Yilun Zhao, Roy Bar-Haim, Arman Cohan, and Michal Shmueli-Scheuer.
\newblock Survey on evaluation of llm-based agents.
\newblock \emph{arXiv preprint arXiv:2503.16416}, 2025.

\bibitem[Yue et~al.(2025)Yue, Zheng, Ni, Wang, Zhang, Tong, Sun, Yu, Zhang, Sun, Su, Chen, and Neubig]{yue2024mmmu}
Xiang Yue, Tianyu Zheng, Yuansheng Ni, Yubo Wang, Kai Zhang, Shengbang Tong, Yuxuan Sun, Botao Yu, Ge Zhang, Huan Sun, Yu Su, Wenhu Chen, and Graham Neubig.
\newblock Mmmu-pro: A more robust multi-discipline multimodal understanding benchmark, 2025.

\bibitem[{Z.AI}(2025)]{glm46}
{Z.AI}.
\newblock Glm-4.6v.
\newblock \url{https://z.ai/blog/glm-4.6v}, 2025.

\bibitem[Zhang et~al.(2025)Zhang, Griffiths, Narasimhan, and Press]{zhang2025videogamebench}
Alex~L Zhang, Thomas~L Griffiths, Karthik~R Narasimhan, and Ofir Press.
\newblock Videogamebench: Can vision-language models complete popular video games?
\newblock \emph{arXiv preprint arXiv:2505.18134}, 2025.

\bibitem[Zhang et~al.(2020)Zhang, Walshe, Liu, Guan, Muller, Whritner, Zhang, Hayhoe, and Ballard]{zhang2019atariheadatarihumaneyetracking}
Ruohan Zhang, Calen Walshe, Zhuode Liu, Lin Guan, Karl Muller, Jake Whritner, Luxin Zhang, Mary Hayhoe, and Dana Ballard.
\newblock Atari-head: Atari human eye-tracking and demonstration dataset.
\newblock In \emph{AAAI}, 2020.

\bibitem[Zhao et~al.(2025)Zhao, Raval, Zhang, Mao, Shangguan, Nikolaidis, Wang, and Seita]{zhao2025manipbenchbenchmarkingvisionlanguagemodels}
Enyu Zhao, Vedant Raval, Hejia Zhang, Jiageng Mao, Zeyu Shangguan, Stefanos Nikolaidis, Yue Wang, and Daniel Seita.
\newblock Manipbench: Benchmarking vision-language models for low-level robot manipulation, 2025.

\bibitem[Zhou et~al.(2023)Zhou, Xu, Zhu, Zhou, Lo, Sridhar, Cheng, Ou, Bisk, Fried, et~al.]{zhou2024webarenarealisticwebenvironment}
Shuyan Zhou, Frank~F Xu, Hao Zhu, Xuhui Zhou, Robert Lo, Abishek Sridhar, Xianyi Cheng, Tianyue Ou, Yonatan Bisk, Daniel Fried, et~al.
\newblock Webarena: A realistic web environment for building autonomous agents.
\newblock \emph{arXiv preprint arXiv:2307.13854}, 2023.

\bibitem[Zhou et~al.(2025)Zhou, Zhu, Wen, Shen, and Xu]{zhou2025vision}
Zhongyi Zhou, Yichen Zhu, Junjie Wen, Chaomin Shen, and Yi Xu.
\newblock Vision-language-action model with open-world embodied reasoning from pretrained knowledge.
\newblock \emph{arXiv preprint arXiv:2505.21906}, 2025.

\bibitem[Zhu et~al.(2023)Zhu, Kapoor, Min, Han, Li, Geng, Neubig, Bisk, Kembhavi, and Weihs]{10204393}
Hao Zhu, Raghav Kapoor, So~Yeon Min, Winson Han, Jiatai Li, Kaiwen Geng, Graham Neubig, Yonatan Bisk, Aniruddha Kembhavi, and Luca Weihs.
\newblock Excalibur: Encouraging and evaluating embodied exploration.
\newblock In \emph{CVPR}, 2023.

\end{thebibliography}
}

\appendix

\section{Dataset Selection and Curation Rationale}
\label{sec:appendix_selection_criteria}

The construction of AgentVQA is guided by a principled selection process to ensure the final benchmark is comprehensive, robust, and effective for evaluating the agentic capabilities of Vision-Language Models (VLMs). Our domains: \webagents, \egocentricvideos, \roboticsdomain, \gamesdomain, and \spatialunderstanding are chosen to represent a broad spectrum of agentic challenges, from digital interaction to embodied physical perception and strategic planning. The following five criteria are central to our selection of the 23 source datasets.

\paragraph{High agentic signal}
A primary requirement is that each dataset's tasks must involve a strong ``agentic signal.'' This means moving beyond passive visual question answering (e.g., ``What color is the car?'') to scenarios that demand active reasoning from an agent's perspective. We prioritize datasets where the core task involves:
\begin{itemize}
    \item Choosing actions or predicting the next optimal step.
    \item Evaluating the consequences of a sequence of actions.
    \item Understanding causality and the relationships between actions and outcomes.
    \item Strategic planning toward a specific goal.
\end{itemize}
This focus ensures that AgentVQA directly probes the decision-making and reasoning capabilities that are fundamental to agentic intelligence.

\paragraph{Comprehensive and distinct domain coverage}
To test for generalist agentic behavior, we select datasets that span diverse domains and modalities. Each of the five chosen domains introduces a unique set of challenges and required skills not fully covered by the others:
\begin{itemize}
    \item \webagents: Tests GUI navigation, grounding, and interaction in a digital environment.
    \item \roboticsdomain \& \egocentricvideos: Evaluate embodied perception, spatial awareness, and understanding of the physical world from a first-person perspective.
    \item \gamesdomain: Focus on strategic decision-making, reward modeling, and long-term planning in rule-based environments.
    \item \spatialunderstanding: Assesses foundational spatial awareness and reasoning, a critical component for any embodied agent.
\end{itemize}
This multi-domain approach prevents the benchmark from being too narrow and allows for a holistic assessment of a VLM's generalization ability.

\paragraph{Offline convertibility}
A crucial practical consideration is the feasibility of converting each dataset into a standardized, offline format. This criterion requires that the source data (e.g., trajectories, videos) be sufficiently structured to allow for an unambiguous conversion into a Multiple-Choice Question (MCQ) format. Every resulting MCQ must have a single, verifiably correct answer to enable scalable, deterministic, and efficient evaluation without relying on resource-intensive online simulators.

\paragraph{Public availability and permissive licensing}
To ensure transparency, reproducibility, and accessibility for the broader research community, all selected datasets must be publicly available. Furthermore, their licenses must permit modification and redistribution as part of a new benchmark, allowing us to legally and ethically create and share AgentVQA.

\paragraph{Sufficient scale}
   The selected datasets need to be large enough to provide a diverse pool from which high-quality samples can be curated. A larger source pool increases the likelihood of finding samples that are both challenging and unambiguous across a range of question types. This criterion ensures that our manual curation process (Section \ref{sec:manual_curation}) has adequate coverage of each domain and dataset.

\section{Source Dataset Overview}
\label{sec:appendix_dataset_overview}

Table~\ref{tab:source_dataset_overview} summarizes the source datasets used in AgentVQA. The input and output columns describe each dataset after conversion into AgentVQA. Every output is represented as a multiple choice question, while the listed output type identifies the meaning of the selected option.

\begin{table*}[p]
    \centering
\caption{Source datasets used in AgentVQA. Descriptions and modalities follow the released AgentVQA data fields and the original dataset definitions.}
\label{tab:source_dataset_overview}
    \footnotesize
    \setlength{\tabcolsep}{4pt}
    \renewcommand{\arraystretch}{1.12}
    \begin{tabularx}{\textwidth}{@{}p{3.1cm}X p{1.45cm}p{2.7cm}@{}}
        \toprule
        Dataset & Description & Input type & Output type \\
        \midrule
        \multicolumn{4}{l}{\webagents} \\
        AitW~\cite{rawles2023androidwildlargescaledataset} & Android device control from a task goal, the current screen, and previous actions. & Image & Next UI action \\
        MONDAY~\cite{jang2025scalablevideotodatasetgenerationcrossplatform} & Android and iOS navigation trajectories extracted from instructional videos. & Image & Next UI action \\
        Screenspot~\cite{cheng2024seeclick} & Grounding an instruction to an actionable element in mobile, desktop, or web interfaces. & Image & Click coordinates \\
        Screenspot-pro~\cite{li2025screenspotproguigroundingprofessional} & Grounding an instruction in high resolution interfaces from professional applications. & Image & Click coordinates \\
        \midrule
        \multicolumn{4}{l}{\roboticsdomain} \\
        ERQA~\cite{comanici2025geminirobotics} & Embodied reasoning about robot scenes, actions, trajectories, and task progress. & Multiple images & Reasoning answer \\
        Robo2VLM~\cite{chen2025robo2vlm} & Visual questions derived from real robot manipulation trajectories to test spatial, goal conditioned, and interaction reasoning. & Image & Task state or relation \\
        Roborefit~\cite{lu2023vlgrasp6dofinteractivegrasp} & Grounding a language specified grasp target in a cluttered indoor scene. & Image & Grasp coordinates \\
        ManipBench~\cite{zhao2025manipbenchbenchmarkingvisionlanguagemodels} & Low level manipulation over rigid, articulated, deformable, tool use, and dynamic tasks. & Image & Manipulation trajectory \\
        Share-Robot~\cite{ji2025robobrainunifiedbrainmodel} & Robot manipulation data for task planning, object affordance, and end effector trajectories. & Multiple images & Task description \\
        \midrule
        \multicolumn{4}{l}{\egocentricvideos} \\
        VSI-Bench~\cite{yang2025thinking} & Spatial reasoning from sequential observations of indoor environments. & Video & Spatial answer \\
        Perception-Test~\cite{patraucean2023perceptiontestdiagnosticbenchmark} & Video perception and reasoning over memory, abstraction, physics, and semantics. & Video & Perception answer \\
        OpenEQA~\cite{majumdar2024openeqa} & Questions about real world environments using an agent's episodic visual memory. & Video & Environment answer \\
        VidEgoThink~\cite{cheng2024videgothinkassessingegocentricvideo} & Egocentric understanding across planning, grounding, and reward modeling. & Video & Video reasoning answer \\
        EgoPlan-Bench~\cite{chen2024egoplanbenchbenchmarkingmultimodallarge} & Planning the next step toward a task goal from egocentric task progress. & Video & Next action \\
        \bottomrule



        \toprule
        Dataset & Description & Input type & Output type \\
        \midrule
        \multicolumn{4}{l}{\gamesdomain} \\
        GameQA~\cite{tong2025gamerlsynthesizingverifiablegame} & Game state reasoning across video games, including perception and spatial reasoning. & Image & Game state answer \\
        Atari~\cite{zhang2019atariheadatarihumaneyetracking} & Atari gameplay demonstrations with human actions and simultaneously recorded gaze. & Image & Gameplay action \\
        VideoGameBunny~\cite{taesiri2024videogamebunnyvisionassistantsvideo} & Understanding video game scenes through captions, questions, and scene attributes. & Image & Game understanding answer \\
        MarioQA~\cite{mun2017marioqaansweringquestionswatching} & Event centric, counting, and state questions about Super Mario gameplay. & Video & Gameplay event answer \\
        \midrule
        \multicolumn{4}{l}{\spatialunderstanding} \\
        SpaCE-10~\cite{gong2025space10comprehensivebenchmarkmultimodal} & Compositional spatial reasoning across multiple scenes, including entity presence. & Multiple images & Spatial description \\
        EmbSpatial-Bench~\cite{du2024embspatialbenchbenchmarkingspatialunderstanding} & Object to object and object to scene spatial relationships for embodied tasks. & Image & Spatial relation \\
        Spatial-MM~\cite{shiri2024empiricalanalysisspatialreasoning} & Spatial reasoning about object locations and relations in visual scenes. & Image & Spatial answer \\
        OmniSpatial~\cite{jia2026omnispatialcomprehensivespatialreasoning} & Spatial reasoning across diverse task types and visual scenes. & Image & Spatial answer \\
        3DSRBench~\cite{ma20253dsrbenchcomprehensive3dspatial} & Reasoning about real world three dimensional locations and spatial relations. & Image & 3D spatial answer \\
        \bottomrule
    \end{tabularx}
    
\end{table*}


\section{Hard Negative Generation Pipeline}
\label{sec:appendix_hard_negatives}

The difficulty of AgentVQA is ensured by a systematic process for generating challenging distractor options, or ``hard negatives.'' This process transforms tasks that are not inherently multiple-choice, such as web navigation trajectories, into a standardized and rigorous MCQ format. Our pipeline creates negatives that are both plausible and unambiguously incorrect, forcing models to demonstrate fine-grained understanding rather than relying on simple heuristics. The process is detailed below.

\paragraph{VLM-assisted generation} To select the generation model, we first conducted a preliminary study on a data subset using Gemini 2.5 Pro, GPT-4o, and Qwen 2.5 VL (7B). After manually comparing the generated samples for both difficulty and accuracy, we selected Gemini 2.5 Pro for its superior performance. We therefore employ a VLM-assisted pipeline centered around \textbf{Gemini 2.5 Pro} to generate hard negatives. For each question, we provide the model with comprehensive context, including the primary image or screenshot, the task prompt, the ground-truth answer, any relevant action history for sequential tasks, and metadata such as image dimensions and coordinate formats (e.g., (x,y) for points, or ($x_1, y_1, x_2, y_2$) for bounding boxes). For grounding tasks, we also visually render the correct action's location onto the image. Given this context, we prompt the model to generate three distinct hard negatives as a combination of two types: \textbf{near-miss negatives} and \textbf{semantically similar negatives}. Near-miss negatives are distractors spatially close to the correct answer that test precision, such as coordinates a few pixels away from a correct UI element. Semantically similar negatives are conceptually related to the correct answer and test the model's ability to differentiate between similar objects, such as choosing a ``Mobile'' phone when the correct answer is a ``Tablet.'' The exact prompt used for generating these ``hard negatives'' is in Appendix ~\ref{sec:appendix_hard_negative_prompt}

\paragraph{Verification and filtering} To ensure quality, the generated negatives undergo a two-stage verification process. First, an automated script filters grounding-based distractors by verifying their coordinates fall outside the ground-truth bounding box, programmatically confirming the option is a ``negative.'' Following this, a subset of these samples undergoes a rigorous manual review. We manually verify each distractor for both plausibility, ensuring the option is a believable choice that could confuse a model, and unambiguous incorrectness, which guarantees there is only one correct answer to the MCQ. This human-in-the-loop step is critical for ensuring the quality of the generated negatives prior to the manual curation stage described in Section \ref{sec:manual_curation}. The hard negatives produced by this pipeline form the initial candidate pool. The final 1000-sample benchmark undergoes additional manual curation (Section \ref{sec:manual_curation}) to verify correctness, difficulty, and the absence of any ambiguity.

\section{\textcolor{black}{Definition of Agentic Tasks}}
\label{agentic}

\textcolor{black}{The community defines agentic tasks as those requiring a system to "observe their environment and act in it in order to achieve goals"~\cite{bengio2025superintelligentagentsposecatastrophic}, encompassing perception to sense the state, intelligence to reason and plan, and affordances to execute actions~\cite{plaat2025agenticlargelanguagemodels}. While traditional image VQA treats visual understanding as an end in itself, agentic VQA frames perception as a means to action. Here, each answer must inform executable decisions rather than merely describe what is seen~\cite{reed2022generalistagent}. To ensure AgentVQA captures this full capability range, we structure our sub-tasks into a unified hierarchy: Action Grounding tasks (e.g., Tap) test the critical translation of semantic intent into executable spatial actions~\cite{reed2022generalistagent}; Spatiotemporal Reasoning tasks evaluate the dynamic mental maps required for movement; State Understanding tasks (e.g., Functional Reasoning) ensure the agent can accurately parse the environment to inform decision-making; and Strategic Planning tasks (e.g., Reward Modeling) assessing long-term outcomes through multi-step reasoning~\cite{sutton1998reinforcement}.}

\section{Answer Extraction Logic}
\label{sec:appendix_extraction_logic}

To ensure consistent and automated evaluation, we employ a multi-strategy answer extraction logic. Although the prompt instructs models to output only a single letter corresponding to their final choice, outputs can vary. Our programmatic parser is designed to robustly handle these variations.

\paragraph{Extraction strategy} The extraction process follows a prioritized sequence of rules. First, our parser attempts a direct match, checking if the model's entire stripped output is exactly a single, valid option letter (e.g., `A'). If this primary strategy fails, it deploys a fallback routine that applies a series of regular expression patterns. These patterns are designed to find common answer declarations such as ``Answer: A,'' or ``Option A.'' We note that these regular expressions are occasionally adapted to accommodate the unique, consistent output formats of specific models. If no unambiguous answer can be identified after applying all strategies, the question is skipped and marked unevaluated to maintain the integrity of the evaluation. The complete prompt is available in Appendix \ref{sec:appendix_eval_prompt}.



\section{\textcolor{black}{Robustness Analysis of MCQ Format}}
\label{app:mcq_robustness}

\textcolor{black}{To ensure the reliability of the multiple-choice format, we conducted extensive ablation studies to quantify the impact of option ordering and rule out random guessing as a confounding factor.}

\textcolor{black}{\paragraph{Robustness to Option Ordering} We evaluated 5 representative models across 6 datasets using 5 distinct, seeded random permutations of option orders. As shown in Table \ref{tab:option_shuffling}, the model performance remains highly consistent across permutations, with a standard deviation of approximately $1.9\%$. This confirms that the rankings reported in AgentVQA are stable artifacts of model capability rather than positional bias.}

\begin{table}[h]
\centering
\caption{\textcolor{black}{Robustness to Option Ordering: Average accuracy across 5 random permutations of option orders. The low standard deviation indicates high stability.}}
\label{tab:option_shuffling}
\small

\begin{tabular}{l|ccccc|c}
\toprule
\textbf{Shuffle Iteration} & \textbf{0} & \textbf{1} & \textbf{2} & \textbf{3} & \textbf{4} & \textbf{Std. Dev} \\
\midrule
Average Accuracy & 56.0\% & 54.8\% & 57.0\% & 51.7\% & 54.5\% & $\sim 1.9\%$ \\
\bottomrule
\end{tabular}
\end{table}

\paragraph{Refuting random guessing} To verify that model scores reflect genuine engagement with the task, we evaluated constant-selection baselines (e.g., ``Select All A'', ``Select All B'') across the benchmark. These strategies yielded accuracies of 28\% (A), 27\% (B), 23\% (C), and 22\% (D), averaging \textbf{25.0\%}, which aligns with the theoretical random baseline for 4-option MCQs. Since all 22 models in Table \ref{tab:my_label_custom_width} achieve overall accuracies above the 25\% random-guessing baseline (lowest: 28\%), we conclude that AgentVQA measures performance beyond chance at the full-benchmark level.


\section{Error Modes}
\label{sec:appendix_error_mode_methodology}

To qualitatively understand model failures, we conduct an in-depth error mode analysis on Gemini 3.1 Pro~\citep{gemini3}. The process involves a three-stage methodology to develop a taxonomy, annotate failures, and verify the results. The resulting distribution of errors is visualized in Figure~\ref{fig:error_modes}.

\paragraph{Taxonomy development and annotation} First, we prompt the model to output a reasoning chain alongside its final answer for every question. We then manually review a diverse, stratified subset of incorrect predictions to identify and define a comprehensive taxonomy of common failure patterns for each of the five agentic domains. With this taxonomy established, we use Gemini 3.1 Pro for large-scale annotation. For each failure, we provide the model with the visual input, all multiple-choice options, the ground-truth answer, the model's incorrect chosen option, the model's full reasoning chain, and any other relevant metadata. We then prompt it to assign the single most fitting error mode from our predefined taxonomy. The prompt used for assigning error modes is available in Appendix~\ref{sec:appendix_error_annotation_prompt}.

\paragraph{Verification} To ensure the quality of the automated annotations, we perform a final verification step. We manually review a randomly selected subset with 100 error mode assignments across all domains. This process confirms that the classifications are accurate and consistent with our established definitions, ensuring the reliability of our qualitative analysis.

\subsection{Error Mode Definitions}
\label{sec:appendix_error_mode_definitions}

\paragraph{\webagents}
Grounding Error occurs when the model understands the goal but fails to correctly locate or ground the corresponding UI element, often confusing it with a visually similar or nearby one. Logical Breakdown describes a failure to correctly process a sequence of steps, leading to a logically inconsistent action based on the current state. Visual Hallucination is when the model's reasoning relies on a UI element or piece of information that is not present in the visual input. Missing Inference occurs when the model fails to deduce an implicit but necessary action, such as scrolling to reveal an element. Action Mismatch happens when the model's reasoning correctly identifies the right action, but its final output selects a different, incorrect option. Goal Misunderstanding is a fundamental error where the model misinterprets or ignores a key part of the user's instruction.

\paragraph{\egocentricvideos}
Motion Misperception occurs when the model fails to correctly interpret a process, motion, or physical interaction as it unfolds over time. Perception Gaps are fundamental failures in seeing, where the model incorrectly reports a basic, static fact like an object's presence or count. Spatial Confusion is a failure to build an accurate 3D mental map from video, resulting in an incorrect understanding of object layout or orientation from a given viewpoint. Incomplete Analysis happens when the model perceives all necessary visual facts but fails to execute the required logical steps to reach the correct conclusion. Task Misinterpretation arises when visual perception is correct, but the model misunderstands the nuance or goal of the prompt.

\paragraph{\roboticsdomain}
Spatial Confusion occurs when the model fails to correctly interpret the spatial relationships, location, or orientation of objects in 3D space from a 2D image. Object Misrecognition happens when the model misidentifies an object or fails to ground it based on all of its required attributes like color or size. Action-Time Disconnect is a failure to understand the dynamics of a scene, such as predicting the outcome of an action or correctly sequencing a task. Selection Error occurs when the model's reasoning arrives at the correct conclusion, but it selects the wrong option in its final output. COT Failure describes cases where the chain-of-thought reasoning process breaks down, failing to handle specific quantitative values or fine-grained distinctions.

\paragraph{\gamesdomain}
Visual Miscomprehension is a foundational failure to correctly perceive the game's visual and spatial information, such as object locations or relationships. Rule Misunderstanding is a logical error where the model does not correctly apply the game's explicit rules or objectives. Tracking Drift is a failure to mentally simulate and track how the game state changes over time. Reasoning Breakdown is a general failure where the model's internal logic collapses under the task's complexity or becomes confused about the objective, leading to a flawed conclusion. Context Hallucination occurs when the model's reasoning becomes detached from the game state, inventing information that does not exist.

\paragraph{\spatialunderstanding}
Viewpoint Gap occurs when the model identifies objects but fails to construct an accurate 3D map of their relationships, leading to errors in judging depth or distance. Categorization Confusion occurs when perception is visually correct, but the model applies the wrong semantic label or category to an object. Detection Hallucination is a fundamental perceptual failure where the model either misses a present object or ``sees'' an object that is not there. Logic Collapse is a failure cascade where the model first fails to perceive an object and then tries to compensate by inventing a logical path forward based on hallucinated information. Scope Misinterpretation happens when the model misunderstands the required granularity or intent of the question, particularly for counting or classification tasks.

\begin{table}[htbp]
    \centering
\caption{Here we list canonical Vision-Language model agentic benchmarks, organized by domain and their offline/online categorization. We \textbf{bold} datasets that are a part of AgentVQA and suffix datasets that have been filtered or filtered and transformed with + or ++ suffixes, respectively.}
\label{tab:agentic_benchmarks_compact}

    \tiny
    \setlength{\tabcolsep}{3pt}
    \renewcommand{\arraystretch}{1.5}
    \begin{adjustbox}{width=0.98\textwidth,center}
    \begin{tabular}{>{\centering\arraybackslash}p{1.4cm}|>{\centering\arraybackslash}p{0.8cm} >{\raggedright\arraybackslash}p{11.8cm}}
    \toprule
    \textbf{Domain} & \textbf{Mode} & \textbf{Datasets} \\
    \midrule
    
    \multirow{2}{*}{\webagents} 
    & Offline & \textbf{AitW}$^{++}$~\cite{rawles2023androidwildlargescaledataset}, \textbf{ScreenSpot}$^{++}$~\cite{cheng2024seeclick}, \textbf{Monday}$^{++}$~\cite{jang2025scalablevideotodatasetgenerationcrossplatform}, \textbf{Screenspot-Pro}$^{++}$~\cite{li2025screenspotproguigroundingprofessional} \\[0.5ex]
    \cline{2-3}
    & Online & WebArena~\cite{zhou2024webarenarealisticwebenvironment}, VisualWebArena~\cite{koh2024visualwebarena}, TheAgentCompany~\cite{xu2025theagentcompanybenchmarkingllmagents}, AndroidWorld~\cite{rawles2024androidworld}, VideoWebArena~\cite{jang2025videowebarenaevaluatinglongcontext}\\[0.5ex]
    \midrule
    
    \multirow{2}{*}{\parbox{1.4cm}{\centering Egocentric\\Videos}} 
    & Offline & \textbf{Perception-Test}$^{+}$~\cite{patraucean2023perceptiontestdiagnosticbenchmark}, \textbf{VSI-Bench}$^{+}$~\cite{yang2025thinking}, \textbf{VidEgoThink}\textbf{}~\cite{cheng2024videgothinkassessingegocentricvideo}, Ego-Exo4D~\cite{grauman2024egoexo4dunderstandingskilledhuman}, \textbf{OpenEQA}\textbf{}~\cite{majumdar2024openeqa}, \textbf{EgoPlan-Bench}\textbf{}~\cite{chen2024egoplanbenchbenchmarkingmultimodallarge}\\[0.5ex]
    \cline{2-3}
    & Online & EgoThink~\cite{cheng2024egothinkevaluatingfirstpersonperspective} \\[0.5ex]
    \midrule
    
    \multirow{2}{*}{\roboticsdomain} 
    & Offline & \textbf{RoboRefit}$^{++}$~\cite{lu2023vlgrasp6dofinteractivegrasp}, \textbf{ERQA}~\cite{comanici2025geminirobotics}, \textbf{Robo2VLM}$^{+}$~\cite{chen2025robo2vlm}, HoloAssist~\cite{wang2023holoassistegocentrichumaninteraction}, X-Embodiment~\cite{embodimentcollaboration2025openxembodimentroboticlearning}, \textbf{ManipBench}$^{+}$~\cite{zhao2025manipbenchbenchmarkingvisionlanguagemodels}, \textbf{ShareRobot}$^{+}$~\cite{ji2025robobrainunifiedbrainmodel} \\[0.5ex]
    \cline{2-3}
    & Online & CALVIN~\cite{mees2022calvinbenchmarklanguageconditionedpolicy}, VIMA-Bench~\cite{jiang2023vimageneralrobotmanipulation}, BEHAVIOR-1K~\cite{li2024behavior1khumancenteredembodiedai}, RoboCasa~\cite{nasiriany2024robocasalargescalesimulationeveryday}, EmbodiedBench~\cite{pmlr-v267-yang25f} \\[0.5ex]
    \midrule
    
    \multirow{2}{*}{\gamesdomain} 
    & Offline & \textbf{Atari}$^{++}$~\cite{zhang2019atariheadatarihumaneyetracking}, \textbf{GameQA}$^{+}$~\cite{tong2025gamerlsynthesizingverifiablegame}, BASALT~\cite{milani2023beddminerlbasaltevaluation}, \textbf{MarioQA}$^{+}$~\cite{mun2017marioqaansweringquestionswatching}, \textbf{VideoGameBunny}$^{+}$~\cite{taesiri2024videogamebunnyvisionassistantsvideo} \\[0.5ex]
    \cline{2-3}
    & Online & Crafter~\cite{hafner2022benchmarkingspectrumagentcapabilities}, MineDojo~\cite{fan2022minedojobuildingopenendedembodied}, VideoGameBench~\cite{zhang2025videogamebench}, BALROG~\cite{paglieri2025balrogbenchmarkingagenticllm} \\[0.5ex]
    \midrule
    
    \multirow{2}{*}{\parbox{1.4cm}{\centering Spatial\\Planning}} 
    & Offline & \textbf{EmbSpatial-Bench}$^{+}$~\cite{du2024embspatialbenchbenchmarkingspatialunderstanding}, \textbf{SpaCE-10}$^{+}$~\cite{gong2025space10comprehensivebenchmarkmultimodal}, \textbf{Spatial-MM}$^{+}$~\cite{shiri2024empiricalanalysisspatialreasoning}, \textbf{OmniSpatial}$^{+}$~\cite{jia2026omnispatialcomprehensivespatialreasoning}, \textbf{3DSRBench}\textbf{}~\cite{ma20253dsrbenchcomprehensive3dspatial} \\[0.5ex]
    \cline{2-3}
    & Online & HabitatChallenge~\cite{savva2019habitatplatformembodiedai}, EXCALIBUR~\cite{10204393} \\[0.5ex]
    
    \bottomrule
    \end{tabular}
    \end{adjustbox}
\end{table}

\section{Limitations of Open-Ended Grounding Metrics}
\label{app:eval_artifacts}
In this section, we visualize why standard open-ended evaluation metrics (Bounding Box IoU and Distance Thresholds) introduce significant noise into agentic evaluation, and how AgentVQA's MCQ format resolves this ambiguity.

As illustrated in Figure \ref{fig:bbox_failures}, open-ended metrics often force a trade-off: loose thresholds or large bounding boxes allow false positives on irregular shapes (like the scissors), while strict thresholds punish valid actions on large elements (like the mouse). AgentVQA circumvents this by transforming the continuous coordinate space into a discrete decision problem, where distractors are programmatically ensured to be incorrect (e.g., outside the ground truth bounds) yet plausible enough to test precision.

\section{Implications and Suggestions for Practitioners}
\label{app:practitioner_guide}
AgentVQA provides unique diagnostic data unavailable in general VQA benchmarks. Based on the granular failure modes identified in our experiments, we offer the following concrete recommendations for future agentic model design and training:

\begin{itemize}
    \item \textbf{Accelerate Model Iteration:} AgentVQA provides a scalable testbed for cross-domain performance. Evaluating complex agentic questions with a reasoning model judge (generating around 2,000 tokens per decision) requires millions of output tokens for a single pass. AgentVQA mitigates these massive computational costs by enabling rapid, deterministic scoring for faster development cycles.
    \item \textbf{Improve Visual Grounding:} Among the analyzed incorrect Web Agents predictions from Gemini 3.1 Pro, 78\% were classified as grounding errors: the model understood the goal but failed to locate the appropriate UI element. This identifies UI localization as a prominent failure mode in the analyzed sample and suggests it as a target for improvement.
    \item \textbf{Adopt Adaptive Inference:} Since reasoning models often underperform on perception tasks, we recommend adaptive architectures that decouple fast, robust perception modules (for spatial awareness) from expensive Chain-of-Thought (for planning) to prevent ``Thinking Loops.''
    \item \textbf{Implement Domain-Specific Curricula:} The lack of correlation between general VQA and agentic performance suggests generic tuning is inadequate. Our granular error profiles suggest a move toward domain-specific curriculum learning, targeting 3D spatial data for \roboticsdomain versus causal reasoning for \gamesdomain.
    \item \textbf{Deploy Offline Safety Gates:} We propose using AgentVQA as a standardized diagnostic filter to detect critical reliability risks such as hallucinations and recursive loops before deploying agents into high-stakes, open-ended environments.
\end{itemize}

\section{Analysis of "Thinking Loops" as a Failure Mode}
\label{app:loops}

During our analysis of different reasoning models, we identified some models like GLM-4.1V-Thinking getting stuck in a repetitive cycle of reasoning without making any forward progress. For instance, when analyzing one model's reasoning process, we observed it getting stuck in the loop shown below:

\begin{lstlisting}
<think> The person in the video is seen placing a small triangle on the table. The next shape that the person places is a large circle. The person then places another triangle next to the circle. The person then places another triangle next to the circle and the triangle. The person then places another triangle next to the circle and the triangle. The person then places another triangle next to the circle and the triangle........
\end{lstlisting}

\paragraph{Error mode annotation} We use Gemini 3.1 Pro to categorize model failures into predefined error modes (Section \ref{sec:appendix_error_mode_methodology}). All automated assignments were manually verified on a random subset of 100 samples.

\section{Evaluation Prompt}
\label{sec:appendix_eval_prompt}

We use a standard, simple prompt structure that provides the model with the question, any relevant context such as action history, and concludes by instructing it to output only the single letter corresponding to its final choice. The sample prompt structure is mentioned below (Action History would be removed in case of non-trajectory based questions). The prompt below illustrates the Web Agents version of the standard evaluation template, which is adapted slightly across domains to accommodate differences in visual inputs, task context, and question types:









\begin{tcolorbox}
\textbf{Evaluation Prompt Template}

\vspace{1em}
This is the standard template used for model evaluation. Placeholders like \texttt{\{episode\_goal\}} are replaced with the specific data for each question.

\hrulefill
\vspace{1em}

\textbf{Goal:} \texttt{\{episode\_goal\}} \\

\textbf{Action History:} \\
\texttt{\{action\_history if action\_history.strip() else "No previous actions"\}} \\

\textbf{Question:} \texttt{\{question\}} \\

\textbf{Options:} \\
\texttt{\{options\_text\}} \\
\vspace{1em}

Based on the provided screenshot, goal, action history, and question, select the single best option from the list above. \\
\textbf{IMPORTANT:} Your response must be EXACTLY one character (A, B, C, or D) with no other text, explanation, or formatting.\\
\\\textbf{Answer:}
\end{tcolorbox}

\newpage
\section{Hard Negative Generation Prompt}
\label{sec:appendix_hard_negative_prompt}

This is the primary prompt template used to generate hard negative distractors for datasets like AITW; we use a similar but tailored schema for other datasets requiring transformation.
\begin{tcolorbox}
\textbf{Context:}
\begin{itemize}
    \item The UI screenshot has dimensions \texttt{\{width\}}x\texttt{\{height\}} pixels.
    \item The image provided has a green dot marking the location of the correct action if it is a tap.
    \item The correct action is: \textbf{\texttt{\{correct\_action\_string\}}}
\end{itemize}

\textbf{Goal:} \texttt{\{goal\}}

\textbf{Previous Actions (History):} \\
\texttt{\{history if history else "This is the first step."\}}

\vspace{1em}
\hrulefill
\vspace{1em}

\textbf{Your Task:} \\
Generate \textbf{three} distinct and plausible but \textbf{incorrect} distractor actions. These distractors should be designed to confuse a tester. Use a mix of the following strategies:
\begin{enumerate}
    \item \textbf{Near-Miss:} An action of the same type as the correct one but slightly off (e.g., tapping right next to the correct button).
    \item \textbf{Semantic-Confusion:} An action on a different but visually or functionally similar element (e.g., tapping 'blacktooth Settings' when 'Wi-Fi Settings' is correct).
\end{enumerate}

\textbf{Supported Action Formats (Use these formats EXACTLY):}
\begin{itemize}
    \item \texttt{Tap: [x, y]} (where x and y are integer pixel coordinates)
    \item \texttt{Swipe: DIRECTION} (where DIRECTION is one of 'Up', 'Down', 'Left', 'Right')
    \item \texttt{Type: 'text to type'} (where 'text to type' is a plausible but incorrect string)
    \item \texttt{Button: ACTION} (where ACTION is one of 'Press Back', 'Press Home', 'Press Enter')
\end{itemize}

\textbf{Important Rules:}
\begin{itemize}
    \item Do \textbf{NOT} generate the correct answer (\texttt{\{correct\_action\_string\}}).
    \item Ensure all tap coordinates are within the image bounds (width: \texttt{\{width\}}, height: \texttt{\{height\}}).
    \item For "Tap" distractors, analyze the image to choose locations that are genuinely incorrect but tempting.
    \item Generate a diverse set of three distractors.
    \item You also need to ensure that for near miss in tap the point is actually a negative.
\end{itemize}

Respond \textbf{ONLY} with a valid JSON object in the following format:
\begin{verbatim}
{"distractors": ["ACTION_TYPE: value", "ACTION_TYPE: value",
"ACTION_TYPE: value"]}
\end{verbatim}
\end{tcolorbox}

\newpage
\section{Error Mode Annotation Prompt}
\label{sec:appendix_error_annotation_prompt}

The following prompt is used to have Gemini 3.1 Pro assign a specific error mode to a model's incorrect prediction; the prompt provides full context about the question and the model's failure, along with a detailed taxonomy of all possible error modes and their definitions. We present the Robotics-specific prompt as an example; analogous prompts tailored to each domain and its corresponding error-mode taxonomy are used for the other four domains.

\begin{tcolorbox}
You are an expert AI agent evaluator specializing in embodied robotics and robotic reasoning tasks. Your task is to analyze why a vision-language model failed on a specific question and categorize the failure.

\hrulefill
\vspace{1em}

\subsection{CONTEXT}

QUESTION\textbf{:} \texttt{\{question\}} \\ \\
QUESTION TYPE: \texttt{\{question\_type\}} \\ \\
NUMBER OF IMAGES: \texttt{\{num\_images\}} \\ \\
DATASET: \texttt{\{dataset\_name\}}\\

\subsection{VISUAL ANALYSIS}
Some description about the expected image scene.

\subsection{FAILURE DETAILS}
MODEL'S PREDICTION: \texttt{\{model\_prediction\}} \\
CORRECT ANSWER: \texttt{\{correct\_answer\}} \\
MODEL'S REASONING: \texttt{\{reasoning (if available)\}}

\subsection*{TASK}
Analyze the model's failure in this embodied robotics reasoning task. Consider whether the model correctly understood the spatial relationships, identified objects and their attributes, predicted action outcomes, handled quantitative aspects, comprehended the instruction, and mapped its reasoning to the correct output. Focus on identifying the root cause using the specialized robotics error modes below.

\subsection*{ERROR MODES}
\textit{\{A detailed list of all error modes and their definitions is provided here.\}}

\vspace{1em}
\hrulefill
\vspace{1em}

Respond with a JSON object containing the 'error\_mode' and a brief 'justification' for your choice. If you think that none of the error\_modes apply then write NaN in error mode and justify in justification and also propose some new error mode in the justification and explain how this one aligns well with this sample as compared to others. It is not always necessary that an error mode applies so you can output NaN along with the justification and closest error mode from list and your idea of error mode for this.
\end{tcolorbox}




\end{document}